%% file: main.tex
\documentclass[nohyperref]{article}

\usepackage[utf8]{inputenc}
\usepackage[hidelinks]{hyperref}
\usepackage{microtype}
\usepackage{graphicx}
\usepackage{booktabs} 

\usepackage{hyperref}

\usepackage[dvipsnames]{xcolor}
\usepackage{times}
\usepackage{soul}
\usepackage{url}

\usepackage[small]{caption}
\usepackage{graphicx}

\usepackage{amsthm}
\usepackage{booktabs}

\usepackage{algorithm}
\usepackage[accepted]{icml2022}

\usepackage[cmex10]{amsmath}
\usepackage{amssymb}
\usepackage{mathtools}
\usepackage{amsthm}

\usepackage[capitalize,noabbrev]{cleveref}

\theoremstyle{plain}
\newtheorem{theorem}{Theorem}[section]

\theoremstyle{definition}
\newtheorem{definition}[theorem]{Definition}

\theoremstyle{remark}

\newcommand{\modelName}{\textsc{VISL}\xspace} 

\usepackage[textsize=tiny]{todonotes}

\icmltitlerunning{Simultaneous Missing Value Imputation and Structure Learning with Groups}

\newcommand{\cz}[1]{{\color{blue} #1}} 

\usepackage{xspace}
\usepackage{hyperref}
\usepackage{tikz}
\usetikzlibrary{positioning, fit, arrows.meta, arrows} 
\usepackage{wrapfig}
\usepackage{booktabs}       
\usepackage{subcaption}
\usepackage{pdflscape}

\definecolor{myblue}{cmyk}{0.849, 0.738, 0.078, 0}
\definecolor{mygreen}{cmyk}{0.616, 0.0725, 0.894, 0}

\usepackage[ruled,vlined,algo2e]{algorithm2e}

\input{definition}

\begin{document}

\twocolumn[
\icmltitle{Simultaneous Missing Value Imputation and Structure Learning with Groups}



\icmlsetsymbol{equal}{*}
\icmlsetsymbol{intern}{$\dagger$}

\begin{icmlauthorlist}
\icmlauthor{Pablo Morales-Alvarez}{intern,gran}
\icmlauthor{Wenbo Gong}{Microsoft}
\icmlauthor{Angus Lamb$^\dagger$}{G}
\icmlauthor{Simon Woodhead}{Eedi}
\icmlauthor{Simon Peyton Jones$^\dagger$}{E}
\icmlauthor{Nick Pawlowski}{Microsoft}
\icmlauthor{Miltiadis Allamanis}{Microsoft,equal}
\icmlauthor{Cheng Zhang}{Microsoft,equal}
\end{icmlauthorlist}

\icmlaffiliation{gran}{University of Granada}
\icmlaffiliation{Microsoft}{Microsoft Research}
\icmlaffiliation{G}{G-Research}
\icmlaffiliation{E}{Epic Games}
\icmlaffiliation{Eedi}{Eedi}

\icmlcorrespondingauthor{Cheng Zhang}{Cheng.Zhang@microsoft.com}

\icmlkeywords{Machine Learning, ICML}

\vskip 0.3in
]



\printAffiliationsAndNotice{\icmlEqualContribution} 

\begin{abstract}
Learning structures between groups of variables from data with missing values is an important task in the real world, yet difficult to solve. One typical scenario is discovering the structure among topics in the education domain to identify learning pathways. Here, the observations are student performances for questions under each topic which contain missing values. 
However, most existing methods focus on learning structures between a few individual variables from the complete data. In this work, we propose \modelName, a novel scalable structure learning approach that can simultaneously infer structures between groups of variables under missing data and perform missing value imputations with deep learning. Particularly, we propose a generative model with a structured latent space and a graph neural network-based architecture, scaling to a large number of variables. Empirically, we conduct extensive experiments on synthetic, semi-synthetic, and real-world education data sets. We show improved performances on both imputation and structure learning accuracy compared to popular and recent approaches. 
\end{abstract}

\input{intro}

\input{method}

\input{related}
\input{exp}
\input{conclusion}

\bibliography{example_paper}
\bibliographystyle{icml2022}

\newpage
\appendix
\onecolumn
\input{appendix}

You can have as much text here as you want. The main body must be at most $8$ pages long.
For the final version, one more page can be added.
If you want, you can use an appendix like this one, even using the one-column format.

\end{document}

%% file: definition.tex
\usepackage{bm}
\usepackage{color}
\usepackage{amssymb}

\usepackage{algorithmic}
\usepackage{algorithm}

\usepackage{multirow}

\def\p{{\mathrm{p}}}
\def\q{{\mathrm{q}}}

\def\btheta{{\boldsymbol{\theta}}}

\def\bmu{{\boldsymbol{\mu}}}

\def\bsigma{{\boldsymbol{\sigma}}}

\def\bchi{{\boldsymbol{\chi}}}

\def\bh{{\mathbf h}}

\def\bx{{\mathbf x}}

\def\bz{{\mathbf z}}

\def\bG{{\mathbf G}}

\def\bI{{\mathbf I}}

\def\bX{{\mathbf X}}

\def\bZ{{\mathbf Z}}

\DeclareMathOperator*{\argmax}{arg\,max}

%% file: intro.tex
\section{Introduction}\label{sec:intro}

Understanding the structural relationships among different variables provides critical insights in many real-world applications, such as medicine, economics and education \citep{sachs2005causal,zhang2013integrated}. However, it is commonly impossible to perform randomized controlled trials for many real-world applications due to ethical or cost considerations.
Thus, learning graphs from observed data, known as structure learning,   
has recently made remarkable progress \citep{fatemi2021slaps,yu2019dag,zheng2018dags,zheng2020learning}.

For many applications, {variables in the data can be gathered into semantically meaningful groups, where useful insights are at group level}. For example, in finance, one may be interested in how a financial situation influences different industries (i.e. groups) instead of individual companies (i.e. variables). Similarly, in education,
the data can contain student responses to thousands of individual questions (i.e. variables), where each question belongs to a broader topic (i.e. groups).
Again, it is insightful to find relationships between topics instead of individual questions.
Moreover, real-world data such as educational data is inherently sparse since it is not feasible to ask every question to every student; the dimensions of the data in {terms} of {the} number of variables {and} the number of observations are very high, posing a scalability challenge. 
Despite the progress in structure learning, no existing method can {discover} group-wise relationships given large-scale partially observed data.

In this work, we present \modelName (missing \underline{v}alue \underline{i}mputation with \underline{s}tructural \underline{l}earning), a novel approach to simultaneously tackle group-wise structure learning and missing value imputations driven by the real-world topic relationship discovery in an education setting.
This is accomplished by combining variational inference with a generative model that leverages a structured latent space and a decoder based on message-passing Graph Neural Networks (GNN) \citep{gilmer2017neural}.
Namely, the structured latent space endows each group of variables with its latent subspace, and the interactions between the subspaces are regulated by a GNN whose behavior depends on the inferred graph from variational inference, see \autoref{fig:vicause}(a).
\modelName satisfies all the desired properties: it leverages continuous optimization of the structure learning to achieve scalability \citep{zheng2018dags, zheng2020learning}; the \modelName formulation naturally handles missing values,
and it can discover relations at different levels of granularity with pre-defined groups.
Empirically, we evaluate \modelName on one synthetic and two real-world problems including the aforementioned education scenario. 
\modelName shows improved performance in both missing data imputation and structure learning accuracy compared to popular and recent approaches for each task. We worked closely with an education domain expert to evaluate the learned topic relationships, and our model has provided insightful results as recognized by the domain experts.

%% file: method.tex
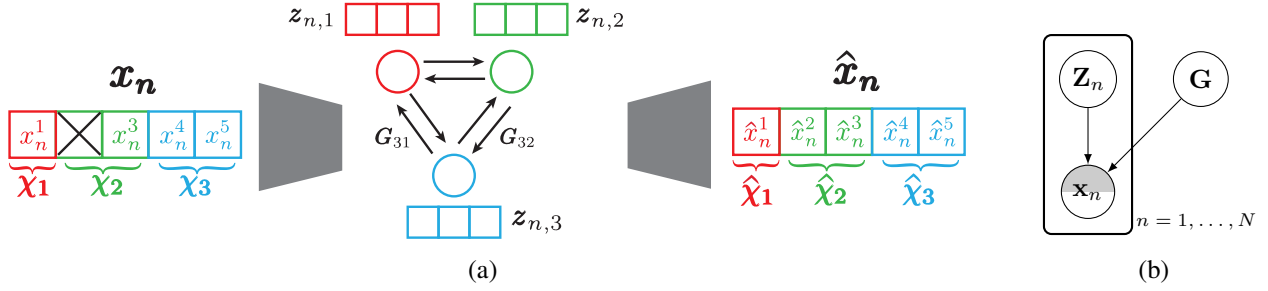
\begin{figure*}
    \centering
    \begin{tabular}{c@{\hskip 1cm}c}
    \includegraphics[scale=0.45]{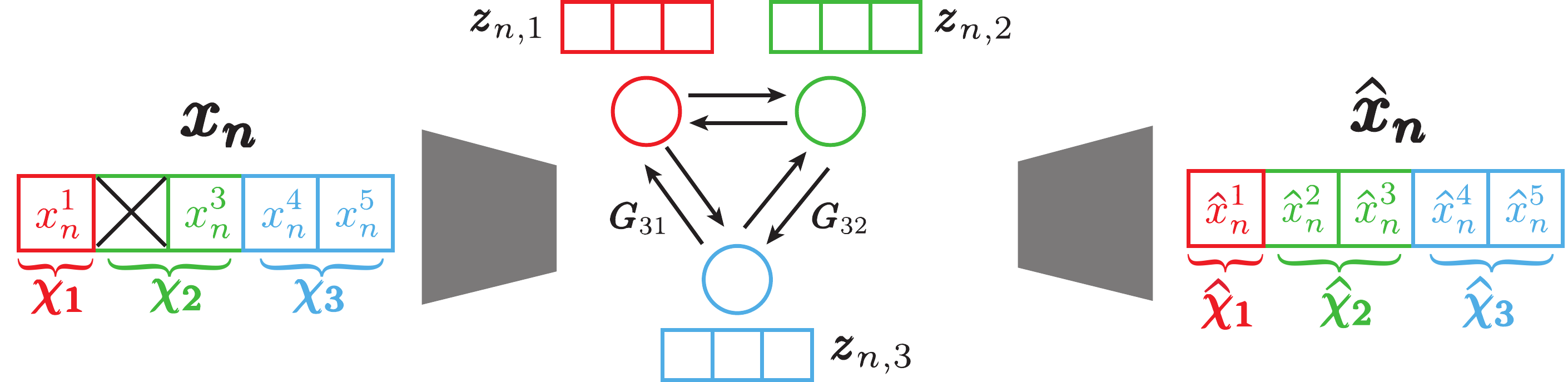}

        &
        \begin{tikzpicture}
\tikzstyle{scalarnode} = [circle, draw, fill=white!11,
text width=1.2em, text badly centered, inner sep=2.5pt]
\tikzstyle{scalarnodetransparent} = [circle, draw,
text width=1.2em, text badly centered, inner sep=2.5pt]
\tikzstyle{arrowline} = [draw,color=black, -latex]
;
\tikzstyle{dasharrowline} = [draw,dashed, color=black, -latex]
;
\tikzstyle{surround} = [thick,draw=black,rounded corners=1mm]
\begin{scope}
    \clip (-.36,0) rectangle (0.36,0.36);
    \draw [fill=black!20,draw=none](0,0) circle(.36);
    \draw [draw=none](-.36,0) -- (.36,0);
\end{scope}
\node [scalarnodetransparent] at (0,0) (x) {$\bx_n$};
\node [scalarnode] at (0, 1.5) (z) {$\bZ_n$};
\node [scalarnode] at (1.5, 1.5) (A) {$\bG$};
\node [] at (1.44, -0.4) (t) {\scriptsize $n=1,\dots,N$};
\path [arrowline] (z) to (x);
\path [arrowline] (A) to (x);

\node[surround, inner sep = .2cm] (f_N) [fit = (z)(x)] {};
\end{tikzpicture}
        \\[1mm]
        (a) &  (b) \\
    \end{tabular}
\vspace{-3mm}
    \caption{(a) Graphic representation of \modelName.{ \modelName is a variational auto-encoder based framework. Observations from each group are encoded into low dimensional latent variables. The structure is treated as {a} global latent variable. {A} GNN based decoder is used to decode the latent variables to observations.}
    (b) Probabilistic graphical model for \modelName, {where the partial observation $\bx$ is generated from its local latent variable $\bz$ and the global latent variable $\bG$ which characterizes the structures. }}
    \label{fig:vicause}
\vspace{-3mm}
\end{figure*}

\section{Model Description}\label{sec:model}

In the following, we present the formulation of \modelName for scalable group-wise structure learning with partial observations using {a} novel deep generative model based framework.   

\subsection{Problem setting}\label{sec:model_notation}
 
Assume a training data set $\bX=\{\bx_n\}_{n=1}^N$ with $\bx_n\in\mathbb{R}^{D}$. The observed and missing values are denoted as $\bX_O$ and $\bX_U$, respectively, where we assume the data are missing completely at random (MCAR) or missing at random (MAR). In \autoref{app: MAR}, we explain how to handle MAR. 
In particular, variables can be gathered into $M$ pre-defined groups, where each can be denoted as $\bchi_{n,m}=[x_{n,i}]_{i\in \mathcal{I}_m}$. $\mathcal{I}_m$ containing the variable indices belonging to group $m$ (e.g., $\mathcal{I}_2=[4,5,6]$ indicates group $2$ includes the $4^{\text{th}}$, $5^{\text{th}}$ and $6^{\text{th}}$ variables). One should note that each $\mathcal{I}_m$ may have varying sizes for different $m$ (i.e.~varying group sizes).
The goal of \modelName is to (i) perform missing value imputation for test samples and (ii) infer structures between groups of variables. 
We use the adjacency matrix $\bG\in[0,1]^{M\times M}$ to represent a graph, where $G_{ij}=1$ or $0$ indicates whether there is a directed edge from $i-$th to $j-$th group or not. In the context of the education domain, the above formulation can be rephrased as follows: variable $\bx_n$ containing the student's responses to a set of questions. $x_{i,j}=1$ represents student $i$ has answered question $j$ correctly.
Groups can be defined as the topic associated with each question. $\mathcal{I}_m$ contains the question IDs that belong to the same topic, and $\bchi_m$ represents a group of responses related to that topic. Clearly, not all students can answer every question. Thus, $\bX_O$, $\bX_U$ represent the existing responses and un-answered questions, respectively. The goal of \modelName is to (i) predict students' responses to un-answered questions, which by itself is important in the education domain \cite{wang2020educational, wang2021results}, and (ii) discover the relationships between topics, which can help education experts optimize the learning experience and the curriculum.
For structure learning, we adopt a Bayesian approach for graphs \citep{heckerman2006bayesian}. Namely, we seek to maximize the posterior probability of $\bG$ given partially observed training data $\bX_O$ within the space of all DAGs: 

\begin{align}
\label{eq:causal_objective}
\bG_\star = & \textstyle\argmax_{\bG\in\textrm{DAGs}}{\color{blue}\p(\bX_O|\bG)\p(\bG)}.
\end{align}

To optimize over the structure with the DAG constraint in \autoref{eq:causal_objective}, we resort to recent continuous optimization techniques \citep{castle, zheng2018dags, zheng2020learning}, where a differentiable measure of 'DAG-ness', $\mathcal{R}(\bG)=\mathrm{tr}(e^{\bG \odot \bG})-D-1$, was proposed and is zero if and only if $\bG$ is a DAG.
To leverage this DAG-ness characterisation, we follow \citet{castle, yu2019dag} and introduce a {\color{ForestGreen}regulariser} based on $\mathcal{R}(\bG)$ to favour the DAG-ness of the solution, i.e.
\begin{equation}\label{eq:causal_objective_with_reg}
   \bG_\star = \textstyle\argmax_{\bG} \left( {\color{blue}\p(\bX_O|\bG)\p(\bG)}-\lambda{\color{ForestGreen}\mathcal{R}(\bG)}\right). 
\end{equation}
 
In the following two sections, we present our detailed formulation, training and imputation algorithms of \modelName, that allows the model to infer the latent structure $\bG$ and impute missing values $\tilde{\bx}_U$ in a test sample $\tilde{\bx}\in\mathbb{R}^D$ based on the observed $\tilde{\bx}_O$.

\subsection{Generative model and variational inference}\label{sec:model_and_inference} 
\begin{wrapfigure}{r}{0.5\columnwidth}
\begin{minipage}{0.5\columnwidth}
\begin{algorithm}[H]
\caption{Generative process}\label{alg:gen_story}
\begin{algorithmic}
\small
\STATE $\bG_{ij} \sim \text{Bernoulli}(p_{ij})$
\FOR{$n\in\{1, 2, \cdots, N\}$}
    \STATE $\bZ_n \sim \mathcal{N}(\mathbf{0}, \sigma_z^2\bI)$
    \STATE $\bx_n \sim \mathcal{N}(f_{\theta}(\bZ_n, \bG),\sigma_x^2\bI)$
\ENDFOR
\end{algorithmic}
\end{algorithm}
\end{minipage}
\end{wrapfigure}

For the generation of observation $\bX$, we adopt the latent variable model of \autoref{fig:vicause}. Particularly, given an inferred graph $G$ and latent $\bZ$, the generative path from $\bZ$ to $\bX$ is provided in \autoref{alg:gen_story}, where we use a graph neural network (GNN) decoder that respects the learned graph structure $G$ and the provided grouping structure. 
%
Then the joint model likelihood is
\begin{gather}\label{eq:full_model}
    \p\left(\bX, \bZ, \bG\right) = \p(\bG)\textstyle\prod_n \p(\bx_n|\bZ_n,\bG)\p(\bZ_n).
\end{gather}

\noindent\textbf{Amortized variational inference.} 
The true posterior distribution over $\bZ$ and $\bG$ in \autoref{eq:full_model} is intractable since we use a complex deep learning architecture. Therefore, we resort to an efficient amortized variational inference as in \citet{kingma2013auto, kingma2019introduction}.  Here, we 
consider a fully factorized variational distribution  $\q(\bZ,\bG)= \q_\phi(\bG)\prod_{n=1}^N \q_\phi(\bZ_n|\bx_n)$, where $\q_\phi(\bZ_n|\bx_n)$ is a Gaussian whose mean and (diagonal) covariance matrix are given by an \emph{encoder}.
For $\q(\bG)$, we consider the product of independent Bernoulli distributions over the edges; that is, the presence of each edge from $i$ to $j$ is associated with a probability $p_{ij}$
to be estimated.
With the above formulation, the evidence lower bound (ELBO) is
\begin{align}\label{eq:ELBO}
\nonumber\textrm{ELBO} &= \textstyle\sum_{n}  \left\{\mathbb{E}_{\q_\phi(\bZ_n|\bx_n)\q(\bG)}[\log\p(\bx_n|\bZ_n,\bG)\right. - \\ & \textrm{KL}[\q_\phi(\bZ_n|\bx_n)||\p(\bZ_n)]]\big\} - \textrm{KL}[\q(\bG)||\p(\bG))]. 
\end{align}\vspace{-12pt}

Next, we explain our choice of the generator (decoder), which uses a GNN over a learned graph $\bG$ to 
model the interactions between latent variables, representing the information about each group. Then, we focus on the inference network (encoder), representing the mapping from the group of observed variables to its corresponding latent representation. 
 
\noindent\textbf{Generator}.
The generator (i.e., decoder) takes $\bZ_n$ and $\bG$ as inputs and outputs the reconstructed $\hat\bx_n=f_{\theta}(\bZ_n,\bG)$, where $\theta$ are the decoder parameters.
In order to respect the pre-defined group structure, as shown in \autoref{fig:vicause}, 
$\bZ_n$ is partitioned into $M$ parts, where $\bz_{n,m}$ represents the latent variable for the group of observations $\bchi_{n,m}$.   
This defines a group-wise structured latent space.  
We adopt a two-step process for the generative path $\bZ_n$ to $\bX_n$: (i) GNN message passing 
with respect to the learned graph $\bG$ between latent $\bz_{n,m}$; (ii) final read-out layer to generate $\bX_n$.
 
\noindent\textbf{GNN message passing in the generator}.
In message passing, the information flows between nodes in $T$ consecutive node-to-edge (n2e) and edge-to-node (e2n) operations \citep{gilmer2017neural}. 
At the $t$-th step, we compute an embedding $\bh^f_{i\to j}$ for each edge $i\to j$, called \emph{forward} embedding, which summarizes the information sent from node $i$ to $j$.
Specifically, the n2e/e2n operations in \modelName are
\begin{align}
\label{eq:n2e}
    \textrm{n2e}: &\quad \bh^{(t),f
    }_{i\to j} =
    \mathrm{MLP}^{f}\!\left(\left[\bz_i^{(t-1)}, \bz_j^{(t-1)}\right]\right),\\\label{eq:e2n}
    \textrm{e2n}: &\quad \bz_i^{(t)} =
    \mathrm{MLP}^{e2n}\left(\textstyle\sum_{k\neq i}\bG_{ki}\cdot
    \bh_{k\to i}^{(t),f}
    \right).
\end{align}
Here, $t$ refers to the $t$-th iteration of message passing (that is, $\bZ^{(0)}=\bZ_n$, notice that we omit subindex $n$ for clarity).
Finally, $\mathrm{MLP}^f$, 
and $\mathrm{MLP}^{e2n}$ are MLPs to be trained.

Interestingly, the message passing updates indicate that the information flows between latent nodes if {a} directed edge {is} specified in graph $\bG$. Hence, the inferred structure $\bG$ directly defines relations for latent space $\bZ$ {which contains the information of pre-defined groups}. 
{We show that }under certain conditions, the inferred graph $G$ also represents the group-wise structure in observational space, and the corresponding model can be reformulated to a general \emph{structural equation model} (SEM) \citep{peters2017elements} (see \autoref{app: respect graph G}).

\noindent\textbf{Read-out layer in the generator}.
After $T$ iterations of GNN message passing, we have $\bZ^{(T)}$.
We then apply a final function that maps $\bZ^{(T)}$ to the reconstructed $\hat\bx$, {which} also respects the pre-defined group structure.
Since the observation $\bx=[\bchi_1,\ldots,\bchi_M]$ may contain $\bchi_m$ with different dimensions, we adopt $M$ different MLPs, one for each group as the final read-out layer, to respect the group structure. Namely,
$\hat{\bx}=(g^1(\bz_1^T),\ldots,g^M(\bz_M^T))^\top,$ where $g^m$ represents the MLP for group $m$. 
Thus, the decoder parameters $\btheta$ include the parameters of the following neural networks: $\textrm{MLP}^f$, 
$\textrm{MLP}^{e2n}$ and {$g^m$ for $m=1,\ldots,M$.

\noindent\textbf{Inference network}.
As in standard VAEs, the encoder maps a sample $\bx_n$ to its latent representation $\bZ_n$. 
As discussed before, $\bZ_n$ is partitioned into $M$ parts, where each $\bz_{n,m}$ contains the information of the observation in group $m$.
{Similar to} the read-out layer, we utilize the $M$ MLPs to map groups of observations to the mean/variance of the latent variables:
\begin{align}\label{eq:encoder_variables}
\bmu_n&=\left(\mu^1_{\phi_{\mu_1}}(\bchi_{n,1}),\dots,\mu^M_{\phi_{\mu_M}}(\bchi_{n,M})\right)^\intercal,\\ \nonumber
\bsigma_n&=\left(\sigma^1_{\phi_{\sigma_1}}(\bchi_{n,1}),\dots,\sigma^M_{\phi_{\sigma_M}}(\bchi_{n,M})\right)^\intercal.
\end{align}}Here, $\mu^m_{\phi_{\mu_m}}$ and $\sigma^m_{\phi_{\sigma_m}}$ are neural networks for group $m$.
When missing values are present, we replace them with a constant as in \citep{nazabal2020handling}. A graphic representation of how the encoder respects the structure of the latent space is shown in the appendix, \autoref{fig:structured_mappings}(b).

\subsection{Training \modelName}\label{sec:model_training}
Given the model described above, we propose the training objective to minimize w.r.t. $\theta$, $\phi$ and $\bG$:
\begin{equation}\label{eq:training_loss}
    \mathcal{L}_{\textrm{\modelName}}(\theta,\phi,\bG) = {\color{blue}-\mathrm{ELBO}} + \lambda {\color{ForestGreen} \mathbb{E}_{\q(\bG)}\left[\mathcal{R}(\bG)\right]},
\end{equation}
where ELBO is given by \autoref{eq:ELBO} and the DAG regulariser $\mathcal{R}(\bG)$ was introduced in \autoref{eq:causal_objective_with_reg} to favor the DAG-ness of learned graph $\bG$.

\noindent\textbf{Evaluating the training loss $\mathcal{L}_{\textrm{\modelName}}$}.
\modelName can work with any type of data. The log-likelihood term ($\log p_\theta(\bx_n|\bZ_n,\bG)$ in \autoref{eq:ELBO}) is defined according to the data type. We use a Gaussian likelihood for continuous variables and a Bernoulli likelihood for binary ones.
For the inference of $\bZ$ and $\bG$, the standard reparametrization trick is used to sample $\bZ_n$ from the Gaussian $\q_\phi(\bZ_n|\bx_n)$ \cite{kingma2013auto,kingma2019introduction}.
To backpropagate the gradients through the discrete variable $\bG$, we resort to the Gumbel-softmax trick to sample from $\q(\bG)$ \cite{jang, maddison2016concrete}.
The $\textrm{KL}[\q_\phi(\bZ_n|\bx_n)||\p(\bZ_n)]$ and $\textrm{KL}[\q(\bG)||\p(\bG))]$ terms can be obtained in closed-form since they are Gaussian distributions and independent Bernoulli distributions over the edges, respectively. 
This formulation brings additional advantages in real-life applications since one can easily incorporate domain knowledge and prior information into the \modelName framework. For example, if the existence of a specific edge is known a priori, the edge probability can be set to 0/1 in the prior distribution.
Finally, the DAG-loss regulariser in \autoref{eq:training_loss} can be computed by evaluating the function $\mathcal{R}$ on a Gumbel-softmax sample from $\q(\bG)$.
To adapt the model to different missing levels in the training data $\bX$, we adopt the \emph{masking} strategy \citep{eddi, gong2019icebreaker}, which drops a random percent of the observed values during training.
The entire training procedure for \modelName is summarised in Algorithm \ref{alg:training}.

\begin{algorithm}[t]
\small
\SetKwInOut{Input}{Input}
\SetKwInOut{Output}{Output}
\Input{Training dataset $\bX$, possibly with missing values.}
\For{each batch of samples $\{\bx_n\}_{n\in B}$}{
Drop a percentage of the data for each sample $\bx_n$.\;

Encode $\bx_n$ through the reparametrization trick to sample $\bZ_n\sim\mathcal{N}(\boldsymbol{\mu}_\phi(\bx_n), \boldsymbol{\sigma}_\phi^2(\bx_n))$ using Eq.\ref{eq:encoder_variables}.\;

Use the Gumbel-softmax to sample $\bG$ from $\q(\bG)$.\;

Use decoder to reconstruct $\hat\bx_n=f_{\theta}(\bZ_n, \bG)$.\;

Calculate the training loss $\mathcal{L}_{\textrm{\modelName}}$ (\autoref{eq:training_loss}).\;

Gradient step w.r.t. $\phi$ (encoder parameters), $\theta$ (decoder parameters) and $\bG$ (posterior edge probabilities).\;
}
\Output{Encoder parameters $\phi$, decoder parameters $\theta$, and posterior probabilities over the edges $\bG$.}
\caption{Training \modelName. \label{alg:training}}
\end{algorithm}

\noindent\textbf{Two-step training}. After training, we  obtain the posterior of the graph $\bG$, which respects the underlying structure of the groups as shown in \autoref{app: respect graph G}. {With the trained network, we can impute missing values in the groups where their ancestors contain some observations 
but if a group has no ancestors no information can be propagated during imputation. After learning the graph structure and to facilitate the imputation task, we introduce
a \emph{backwards} edge: for an edge from $j$ to $i$ we denote the backwards edge information as $\bh^b_{i\to j}$ which codifies the information that the $i\to j$ edge lets flow from the $j$-th to the $i$-th node. It is defined in the same way as \autoref{eq:n2e}, i.e.,:   $\bh^{(t),b
    }_{i\to j} =
    \mathrm{MLP}^{b}\!\left(\left[\bz_i^{(t-1)}, \bz_j^{(t-1)}\right]\right)$, where $\mathrm{MLP}^b$ is the backward MLP; and the e2n update (Eq.\ref{eq:e2n}) is modified to $ \bz_i^{(t)} =
    \mathrm{MLP}^{e2n}\left(\textstyle\sum_{k\neq i}\bG_{ki}\cdot \left\{
    \bh_{k\to i}^{(t),f} +\bh_{i\to k}^{(t),b}\right\} \right)$.}

{In summary, }we propose a two-stage training process, where the first stage --- described in previous sections --- focuses on discovering the edge directions between nodes without the $\mathrm{MLP}^b$ (i.e., we do not train the $\mathrm{MLP}^b$).
In the second stage, we fix the graph structure $\bG$ and continue to train the model with the backward MLP.
This two-stage training process allows \modelName to leverage the backward MLP for the imputation task without {updating the graph structure}. 

\noindent\textbf{Revisiting the learning objectives}.
The optimal graph of relationships, denoted as $\bG_\star$ in \autoref{eq:causal_objective_with_reg}, is given by the estimated posterior probabilities of graph $\bG$. 
In addition, the regularizer $\mathcal{R}(\bG)$ provides a way to evaluate if the resulting graph is a DAG. By tuning the regularizer strength $\lambda$, one can ensure that the resulting $\bG^*$ represents a proper DAG.

For imputation, similar to \citet{eddi, nazabal2020handling}, the trained model can impute missing values for a test instance $\widetilde\bx$ as
\begin{equation}\label{eq:imputation}
    \p(\widetilde\bx_U|\widetilde\bx_O, \bX)=
    \mathbb{E}_{\q_\phi(\bZ|\widetilde\bx)\q(\bG)}\p(\widetilde\bx_U|\bZ,\bG).
\end{equation}
Therefore, the distribution over $\widetilde\bx_U$ (missing values) is obtained by applying the encoder and decoder with $\widetilde\bx$ as input.
One important distinction of \modelName compared to \citet{eddi,nazabal2020handling} is that it incorporates the learned structure $\bG$ into the imputation, which helps the model avoid over-fitting due to spurious correlations \citep{castle}.

\noindent\textbf{Special case: variable-wise relations}.
In the above formulation, we have defined \modelName for group-wise structure learning. Variable-wise relations can be regarded as a special case. In particular, we can set $M=D$ and $\mathcal{I}_m=\{m\}$ (see \autoref{fig:structured_latent_space} (a) in the appendix), i.e. each group only contains a single variable. Through this modification, we can further simplify the encoder and read-out layer. Instead of using $M$ different MLPs, a single MLP can be shared across all variables since each group has dimension of 1. The mean function for the encoder is then defined as
\begin{equation}
    \boldsymbol{\mu}_n=\left(\mu_\phi(x_{n,1}),\ldots,\mu_\phi(x_{n,D})\right).
\end{equation}
One can define encoder variance $\boldsymbol{\sigma}$ (\autoref{fig:structured_mappings} (a) in the appendix) and the read-out layer $g$ analogously.

%% file: related.tex
\section{Related Work}

Since \modelName simultaneously tackles missing value imputation and structure learning, we review both fields.
Moreover, we review recent works that utilize structure learning to improve the performance of another deep learning task, similar to \modelName. {Finally, as one of the focused applications of this work is in the education domain, we review recent advances of AI in education.}

\noindent\textbf{Structure learning}. 
Structure learning aims to infer the underlying structures associated with some observations. There are mainly three types of methods: constrained-based, score-based, and hybrid.
Constraint-based ones exploit (conditional) independence tests to find the underlying structure, such as PC \cite{spirtes1991algorithm} and Fast Causal Inference (FCI) \cite{spirtes2000causation}. 
They have recently been extended to handle partially observed data through test-wise deletion and adjustments \cite{strobl2018fast, mvpc}.
Score-based methods find the structure by optimizing a proper scoring function. The core difficulty lies in the number of possible graphs growing super-exponentially with the number of nodes \citep{chickering2004large}. Thus, explicitly solving the optimization can only be done up to a few nodes \citep{ott2003finding,singh2005finding,cussens2017polyhedral}, resulting in significant limitations in scalability. Therefore, approximation methods have been proposed to ease the computational burden, including searching over topological ordering \citep{teyssier2012ordering,scanagatta2015learning,scanagatta2016learning}, greedy search \citep{chickering2002optimal,ramsey2017million}, coordinate descent \citep{fu2013learning,aragam2015concave,gu2019penalized}.

Recently, continuous optimization of structures, called \emph{Notears}, has become very popular within score-based methods \citep{zheng2018dags}. \emph{Notears} proposed a differentiable algebraic characterization of the DAG, allowing an equality-constrained optimization problem to learn the model parameters and graph structures jointly. \emph{Notears} has inspired the development of other methods, \emph{Notears-MLP} and \emph{Notears-Sob} \citep{zheng2020learning}, \emph{Grandag} \citep{lachapelle2019gradient}, and \emph{DAG-GNN} \citep{yu2019dag}, which extend the original formulation to model nonlinear relationships between variables.
However, their formulations cannot handle missing values and have been observed to be sensitive to data scaling \citep{kaiser2021unsuitability}.
In particular, \emph{DAG-GNN} also adopts a specially-designed GNN to perform structure learning \citep{yu2019dag}.
Compared to our formulation, there are three key distinctions: {(i) our model is designed to discover the group-wise relationship, while DAG-GNN and other structured discovery methods focus on variables level structure learning; (ii)} our model is capable of performing missing value imputation and group-wise structure learning simultaneously, whereas the original formulation of DAG-GNN {and related work} can only handle complete data; 
(iii) \modelName adopts Bayesian learning for the underlying graphs, whereas DAG-GNN uses a point estimation. 

\noindent\textbf{Structure deep learning}. 
Continuous optimization for learning structures has been used to boost performance in classification. In CASTLE \cite{castle}, structure learning is introduced as a regulariser for a deep learning classification model. This regulariser reconstructs only the most relevant causal features, leading to improved out-of-sample predictions.
In SLAPS \cite{fatemi2021slaps}, the classification objective is supplemented with a self-supervised task that learns a graph of interactions between variables through a GNN.
However, these works focused on the supervised classification task, and they did not advance the performance of the structure learning itself.

\noindent\textbf{Missing values imputation}.
The relevance of missing data in real-world problems has motivated a long history of research \citep{dempster1977maximum, rubin1976inference}.
A popular approach for this task is to estimate the missing values based on the observed ones through different techniques \citep{Scheffer02dealingwith}.
Here, we find popular methods such as missforest \citep{stekhoven2012missforest}, which relies on Random Forest, and MICE \citep{buuren2010mice}, which is based on Bayesian Ridge Regression.
Also, the efficiency of amortized inference in generative models has motivated its use for missing values imputation. 
This is explored in \citet{wu2018conditional}, although fully observed training data is required. 
This limitation is addressed in both \citet{nazabal2020handling}, where a zero-imputation strategy is used for partially observed data, and \citet{eddi}, where a permutation invariant set encoder is utilized to handle missing values.
\modelName also leverages amortized inference, although the discovered relationships inform the imputation through a GNN.

\noindent\textbf{AI in education}.
Recently, there has been tremendous progress in using AI for educational applications. For example, knowledge training \citep{lan2014time,vie2019knowledge,naito2018predictive}, which focuses on tracking the evolution of the knowledge of some students; grading students' performance \citep{waters2015bayesrank}; generating feedback for students working on coding challenges \citep{wu2019zero}. In particular, most related to \modelName is work on imputing missing values in students' responses to questions. \citet{wang2020educational} adopts a partial VAE \citep{eddi} to perform missing value imputation and personalization. However, partial VAE does not consider the structural relations between questions/topics and cannot perform structure learning. With the additional insights from structure learning, \modelName can provide more information to teachers to help curriculum design than just imputations.

%% file: exp.tex
\begin{figure*}[h]
    \begin{minipage}[b]{0.55\textwidth}
    \centering
    \begin{tabular}{cc}

\begin{tikzpicture}[scale=0.75]
\tikzstyle{scalarnode} = [circle, draw, fill=white!11,
text width=1.2em, text badly centered, inner sep=2.5pt]
\tikzstyle{arrowline} = [draw,color=black, -latex]
\tikzstyle{dasheddouble} = [draw, dashed, color=black, latex'-latex']
\tikzstyle{dasharrowline} = [draw,dashed, color=black, -latex]
\node [scalarnode] at (0,0) (1) {1};
\node [scalarnode] at (-1.5,-1) (2) {2};
\node [scalarnode] at (1.5,-1) (5) {5};
\node [scalarnode] at (-1,-2.5) (3) {3};
\node [scalarnode] at (1,-2.5) (4) {4};

\path [arrowline] (1) to (3);
\path [arrowline] (1) to (4);
\path [arrowline] (1) to (5);
\path [arrowline] (2) to (3);
\path [arrowline] (2) to (4);
\path [arrowline] (2) to (5);
\path [arrowline] (3) to (4);

\end{tikzpicture}

&

\begin{tikzpicture}[scale=0.75]
\tikzstyle{scalarnode} = [circle, draw, fill=white!11,
text width=1.2em, text badly centered, inner sep=2.5pt]
\tikzstyle{arrowline} = [draw,color=black, -latex]
\tikzstyle{dasheddouble} = [draw, dashed, color=black, latex-latex]
\tikzstyle{dasharrowline} = [draw,dashed, color=black, -latex]
\node [scalarnode] at (0,0) (1) {1};
\node [scalarnode] at (-1.5,-1) (2) {2};
\node [scalarnode] at (1.5,-1) (5) {5};
\node [scalarnode] at (-1,-2.5) (3) {3};
\node [scalarnode] at (1,-2.5) (4) {4};

\path [arrowline] (1) to (3);
\path [arrowline] (1) to (4);
\path [arrowline] (1) to (5);
\path [arrowline] (2) to (3);
\path [arrowline] (2) to (4);
\path [arrowline] (2) to (5);
\path [arrowline] (3) to (4);
\path[arrowline, dashed] (0.6,-2.6) -- (-0.6,-2.6);
\path[dasheddouble] (3) -- (5);
\path [dasheddouble] (5) to (4);

\end{tikzpicture}
\\
(a) & (b)
\end{tabular}
\vspace{-3mm}
\captionof{figure}{(a): Structure simulated for one of the synthetic datasets with 5 variables. (b): Graph predicted by \modelName (when the one on the left is used as the true one). \modelName predicts all the true relationships plus some additional ones (dashed edges).
\label{fig:synthetic_graph}}
    \end{minipage}\hfill
    \begin{minipage}[b]{0.41\textwidth}
    \footnotesize
    \centering
    \includegraphics[width=0.45\textwidth]{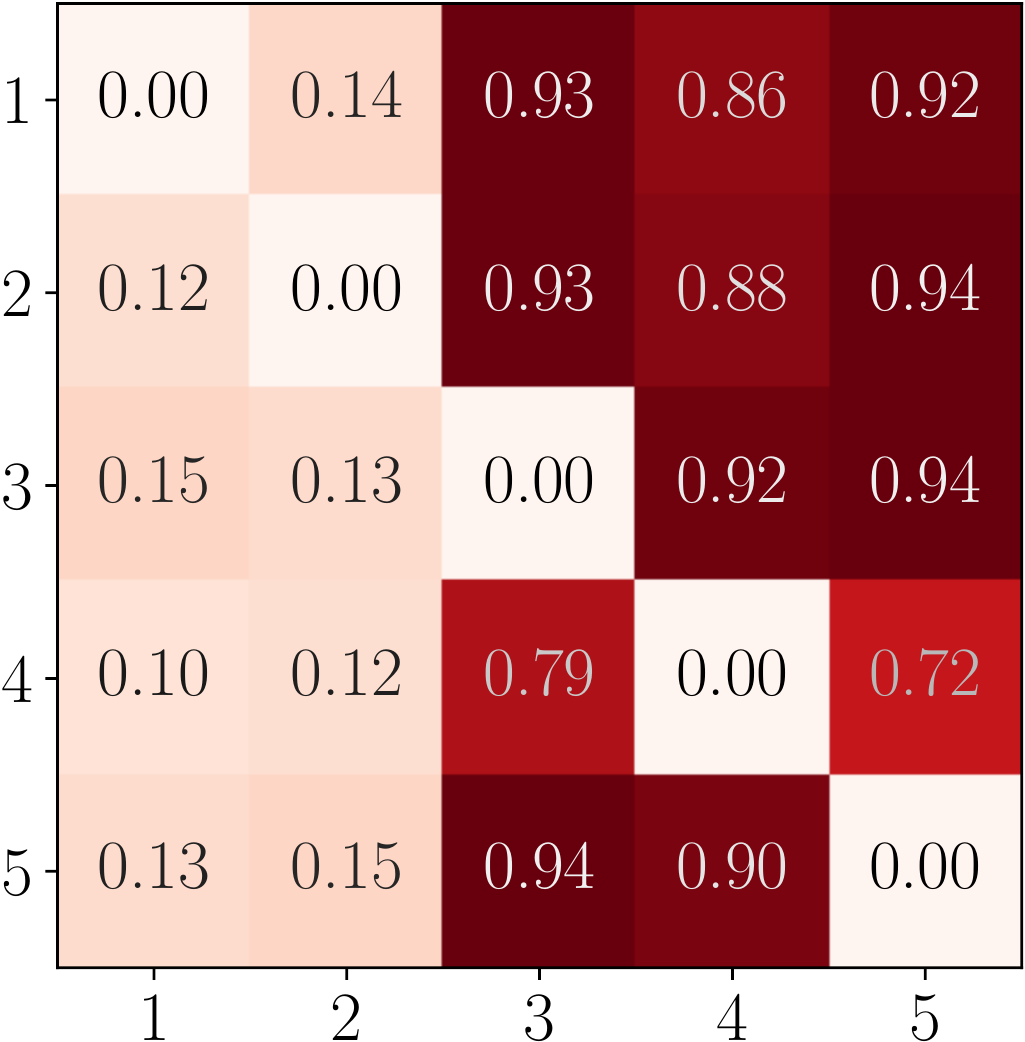}
    \vspace{-2mm}
    \captionof{figure}{Probability of edges obtained by \modelName in the synthetic experiment. By using a 0.5 threshold, we get the predicted graph in \autoref{fig:synthetic_graph}(b). Item $(i,j)$ refers to the probability of edge $i\to j$.
    \label{tab:synthetic_prob_edge}}
    \end{minipage}
    \vspace{-3mm}
\end{figure*}

\begin{table*}[h]
    \begin{minipage}[c]{0.3\textwidth}
    \centering
    \tiny
    \begin{tabular}{rc}
\toprule
 & RMSE \\
\midrule
Majority vote &    0.5442$\pm$0.0032 \\
Mean imputing &    0.2206$\pm$0.0061 \\
MICE          &    0.1361$\pm$0.0046 \\
Missforest    &    0.1313$\pm$0.0025 \\
PVAE  & 0.1407$\pm$0.0043 \\
\modelName &  \textbf{0.1196$\pm$0.0024} \\
\bottomrule
\end{tabular}
\vspace{-3mm}
\captionof{table}{Imputation results for the synthetic experiment. Mean and standard error over 15 datasets.\label{tab:synthetic_imputation}}
    \end{minipage}\hfill
    \begin{minipage}[c]{0.67\textwidth}
    \tiny
    \setlength{\tabcolsep}{4pt}
    \begin{tabular}{rccccccc}
\toprule
{} &  \multicolumn{3}{c}{Adjacency} & \multicolumn{3}{c}{Orientation} & \multirow{2}{*}[-0.2em]{\shortstack{Causal\\accuracy}} \\ 
\cmidrule[0.5pt](lr){2-4}
\cmidrule[0.5pt](lr){5-7}
{} &               Recall &              Precision &               F$_{1}$-score &               Recall &  Precision &               F$_{1}$-score &               {} \\
\midrule
PC &  0.422$\pm$0.056 &  0.634$\pm$0.067 &  0.495$\pm$0.056 &  0.218$\pm$0.046 &  0.328$\pm$0.061 &  0.257$\pm$0.051 &   0.33$\pm$0.046 \\
GES &  0.452$\pm$0.044 &  0.569$\pm$0.036 &  0.491$\pm$0.038 &  0.249$\pm$0.046 &  0.305$\pm$0.053 &   0.270$\pm$0.049 &  0.364$\pm$0.045 \\
NOTEARS (L) &  0.193$\pm$0.028 &  0.443$\pm$0.059 &  0.265$\pm$0.036 &  0.149$\pm$0.023 &  0.367$\pm$0.060 &  0.209$\pm$0.032 &  0.149$\pm$0.023 \\
NOTEARS (NL) &  0.328$\pm$0.039 &  0.489$\pm$0.051 &  0.387$\pm$0.044 &  0.277$\pm$0.032 &  \textbf{0.417$\pm$0.043} &  0.327$\pm$0.035 &  0.277$\pm$0.032 \\
DAG-GNN &  0.443$\pm$0.064 &  0.509$\pm$0.062 &  0.464$\pm$0.061 &  0.352$\pm$0.050 &  0.415$\pm$0.052 &  0.373$\pm$0.049 &  0.352$\pm$0.050 \\
\modelName &   \textbf{0.843$\pm$0.043} &  \textbf{0.679$\pm$0.037} &   \textbf{0.740$\pm$0.033} &   \textbf{0.520$\pm$0.067} &  0.414$\pm$0.058 &   \textbf{0.454$\pm$0.060} &  \textbf{0.726$\pm$0.069} \\
\bottomrule
\end{tabular}
\vspace{-3mm}
    \captionof{table}{Causal discovery results for synthetic experiment (mean and std error over 15 datasets).\label{tab:synthetic_causality}}
    \end{minipage}
\vspace{-5mm}
\end{table*}
\begin{table*}[t]  
\centering
    \scriptsize
    \setlength{\tabcolsep}{3pt}
    \begin{tabular}{rccccccc}
\toprule
{} &  \multicolumn{3}{c}{Adjacency} & \multicolumn{3}{c}{Orientation} & \multirow{2}{*}[-0.2em]{\shortstack{Causal\\Accuracy}} \\ 
\cmidrule[0.5pt](lr){2-4}
\cmidrule[0.5pt](lr){5-7}
{} &               Recall &              Precision &               F$_{1}$-score &               Recall &  Precision &               F$_{1}$-score &               {} \\
\midrule
PC &   0.046$\pm$0.001 &  0.375$\pm$0.006 &  0.082$\pm$0.001 &  0.024$\pm$0.001 &  0.199$\pm$0.011 &   0.044$\pm$0.002 &  0.058$\pm$0.003 \\
GES &  0.110$\pm$0.001 &  0.436$\pm$0.008 &  0.176$\pm$0.002 &  0.082$\pm$0.001 &  0.323$\pm$0.009 &  0.131$\pm$0.003 &   0.121$\pm$0.001 \\
NOTEARS (L) &  0.006$\pm$0.000 &  0.011$\pm$0.001 &  0.008$\pm$0.000 &  0.001$\pm$0.000 &  0.001$\pm$0.001 &  0.001$\pm$0.000 &  0.001$\pm$0.000 \\
NOTEARS (NL) &  0.011$\pm$0.001 &  \textbf{0.644$\pm$0.025} &  0.022$\pm$0.002 &  0.006$\pm$0.001 &  0.354$\pm$0.018 &  0.012$\pm$0.001 &  0.006$\pm$0.001 \\
DAG-GNN &  0.129$\pm$0.028 &  0.272$\pm$0.101 &  0.128$\pm$0.027 &  0.051$\pm$0.010 &  0.126$\pm$0.059 &  0.050$\pm$0.007 &  0.051$\pm$0.010 \\
\modelName & \textbf{0.261$\pm$0.006} &  0.637$\pm$0.009 &  \textbf{0.370$\pm$0.005} &  \textbf{0.236$\pm$0.007} &  \textbf{0.573$\pm$0.005} &  \textbf{0.334$\pm$0.006} &  \textbf{0.245$\pm$0.006} \\
\bottomrule
\end{tabular}
\vspace{-3mm}
    \captionof{table}{Structure discovery results for neuropathic pain data (mean and std error over five runs).\label{tab:neuropathic_causality}}
    \vspace{-5mm}
\end{table*}
\vspace{-2mm}
\section{Experiments}\label{sec:exp}

We evaluate the performance of \modelName in three different problems: a synthetic experiment where the data generation process is controlled, a semi-synthetic problem (simulated data from a real-world problem) with many more variables (Neuropathic Pain), and the real-world problem that motivated the development of the group-level structure learning (Eedi
). {To compare the model performance with related work, the first two datasets are on the variable level, which means that each group only has one variable. In the education setting, we focus on the real-world usage of the method and have worked closely with the domain expert to evaluate the results. } Additional experiments are presented in the appendix.

\textbf{Baselines}.
We consider five baselines for the structure discovery task at the variable level.
PC \cite{spirtes2000causation} and GES \cite{chickering2002optimal} are the most popular methods in constraint-based and score-based approaches, respectively. 
We also consider three recent algorithms based on continuous optimization and deep learning: NOTEARS \citep{zheng2018dags}, the non-linear (NL) extension of NOTEARS \citep{zheng2020learning}, and DAG-GNN \citep{yu2019dag}.
Unlike \modelName, these 
baselines cannot deal with missing values in the training data. Therefore, we work with fully observed training data in the first two sections when using these baselines.
In contrast, the real-world data in the last section comes with partially observed training data, and the goal is to discover group-wise relationships. {These baselines are not applicable.} 
For the missing data imputation task, we also consider five baselines. 
Mean Imputing and Majority Vote are popular techniques used as references, Missforest \citep{stekhoven2012missforest} and MICE \citep{buuren2010mice} are two of the most widely-used imputation algorithms, and PVAE \cite{eddi} is a recent algorithm based on amortized inference.

\textbf{Metrics}.
Imputation performance is evaluated with standard metrics such as RMSE (continuous variables) and accuracy ({binary} variables). For {binary} variables 
, we also provide the area under the ROC and the Precision-Recall curves (AUROC and AUPR, respectively), which are especially useful for imbalanced data (such as Neuropathic Pain).
We follow common practice \citep{glymour2019review, mvpc} regarding structure discovery performance, and consider metrics on the \emph{adjacency} and \emph{orientation}. 
While the former does not take into account the direction of the edges, the latter does.
For both adjacency and orientation, we compute recall, precision and F$_1$-score.
We also provide \emph{causal accuracy}, 
a popular structure discovery metric that considers edge orientation \citep{claassen2012bayesian}.

\subsection{Synthetic experiment}\label{sec:exp_synthetic}

We simulate fifteen synthetic datasets. 
For each simulated dataset, we first sample the true structure $\bG$; see \autoref{fig:synthetic_graph}(a) for an example.
We obtain the samples of the datasets by computing each variable based on its parents using a non-linear mapping based on the $\sin$ function. 
The appendix provides further details, including a visualisation of the generated data in \autoref{fig:synthetic_data}.
For each dataset, we simulate $5000$ training and $1000$ test samples.

\textbf{Imputation performance}. 
\modelName outperforms the baselines in terms of imputation across all synthetic datasets ( \autoref{tab:synthetic_imputation}).
The results grouped by the number of variables are presented by \autoref{tab:synthetic_imputation_extended} in the appendix.
Therefore, \modelName exploits the learned graph to improve imputation by avoiding spurious correlations.

\textbf{Structure discovery performance}.
\modelName obtains better performance than the 
baselines, see \autoref{tab:synthetic_causality}.
The results split by the number of variables are shown in the appendix, \autoref{tab:synthetic_causality_extended}.
Notice that NOTEARS (NL) is slightly better in terms of orientation precision. 
However, this is at the expense of a significantly lower capacity to detect true edges; see the recall and the trade-off between both (F$_1$-score).
In this small synthetic experiment, it is possible to visually inspect the predicted graph. 
\autoref{tab:synthetic_prob_edge} shows the posterior probability of each edge (i.e. the estimated matrix $\bG$) for the simulated dataset that uses the true graph in \autoref{fig:synthetic_graph}(a).
Using a threshold of 0.5, we obtain the predicted graph in \autoref{fig:synthetic_graph}(b).
We observe that all the true edges are captured by \modelName, with some additional edges {due to finite data and non-convex optimization}.

Finally, \modelName can scale to large data both in terms of data points (which benefits naturally from its SGD-based optimization) and dimensionality (thanks to the continuous optimization over the graph space). We demonstrate the computational efficiency {with synthetic data ranging from 4 nodes to 512 nodes} in the appendix, \autoref{tab:synthetic_times}. 

\subsection{Neuropathic pain dataset}\label{sec:exp_neuropathic}
\begin{table}[h]
    \centering
    \scriptsize
    \begin{tabular}{r@{\hskip 2mm}c@{\hskip 2mm}c@{\hskip 2mm}c}
\toprule
{} &               Accuracy &               AUROC & AUPR               \\
\midrule
Majority vote & 0.9268$\pm$0.0003 & 0.5304$\pm$0.0003 & 0.3366$\pm$0.0025 \\
Mean imputing & 0.9268$\pm$0.0003 & 0.8529$\pm$0.0012 & 0.3262$\pm$0.0034 \\
MICE & 0.9469$\pm$0.0007 & 0.9319$\pm$0.0010 & 0.6513$\pm$0.0046 \\

Missforest & 0.9305$\pm$0.0004 & 0.8915$\pm$0.0093 & 0.5227$\pm$0.0033 \\
PVAE &  0.9415$\pm$0.0003 &   0.9270$\pm$0.0007 &  0.5934$\pm$0.0046 \\
\modelName &  \textbf{0.9471$\pm$0.0006} &  \textbf{0.9392$\pm$0.0008} &  \textbf{0.6597$\pm$0.0053} \\
\bottomrule
\end{tabular}
\vspace{-3mm}
\captionof{table}{Imputation results for neuropathic pain data (mean and std error over five runs).\label{tab:neuropathic_imputation}}
 \end{table}

{We evaluate our method using a machine learning benchmark in healthcare applications \citep{tu2019neuropathic}. The dataset contains records of patients regarding the symptoms associated with neuropathic pain. There are 222 variables in this dataset. 
Unlike the previous experiment with continuous data, this dataset has binary variables indicating the symptoms. }
The train and test sets have $1000$ and $500$ patients, respectively. 

\textbf{Imputation performance}.
\modelName shows competitive or superior performance when compared to the baselines, see \autoref{tab:neuropathic_imputation}.
Notice that AUROC and AUPR allow for an appropriate threshold-free assessment in this imbalanced scenario. Indeed, as expected from medical data, the minority of values are 1 (symptoms); here, the prevalence of symptoms is around $8\%$ in the test set.
Interestingly, it is precisely in AUPR where the differences between \modelName and the rest of the baselines are larger except MICE, whose performance is very similar to that of \modelName in this dataset.

\textbf{Structure discovery results}.
As in the synthetic experiment, \modelName outperforms the causality-based baselines; see \autoref{tab:neuropathic_causality}. 
Notice that NOTEARS (NL) is slightly better in terms of adjacency-precision, i.e. the edges that it predicts are slightly more reliable.
However, this is at the expense of a significantly lower capacity to detect true edges, see the recall and the trade-off between both (F$_{1}$-score).


\subsection{Eedi topics dataset}\label{sec:exp_eedi}

\begin{figure*}
    \centering
    \begin{tikzpicture}
\tikzstyle{square} = [rectangle, draw, fill=white!11,
text width=1.2em, text badly centered, inner sep=2.5pt]
\tikzstyle{box} = [rectangle,thick,draw=blue!75,fill=blue!20,minimum size=2mm,rounded corners=.2ex]
\tikzstyle{arrowline} = [draw,color=black, -latex]
\tikzstyle{surround} = [thick,draw=black,rounded corners=1mm]
\node [box] at (-0.5,0) (Maths) {\scriptsize Maths};

\node [box] at (-4.9,-1) (Algebra) {\scriptsize Algebra};
\node [box] at (-0.5,-1) (Number) {\scriptsize Number};
\node [box] at (3.3,-1) (Geometry) {\scriptsize Geometry and measure};

\node [box] at (-6.8,-2) (A_1) {\scriptsize...};
\node [box] at (-4.9,-2) (A_2) {\scriptsize Solving equations};
\node [box] at (-3,-2) (A_3) {\scriptsize...};
\node [box] at (-2.3,-2) (N_1) {\scriptsize...};
\node [box] at (-0.5,-2) (N_2) {\scriptsize Negative numbers};
\node [box] at (1.3,-2) (N_3) {\scriptsize...};
\node [box] at (2.3,-2) (G_1) {\scriptsize...};
\node [box] at (3.3,-2) (G_2) {\scriptsize Angles};
\node [box] at (4.3,-2) (G_3) {\scriptsize...};

\node [box] at (-6.9,-3) (AA_1) {\scriptsize...};
\node [box] at (-4.9,-3) (AA_2) {\scriptsize Quadratic equations};
\node [box] at (-2.9,-3) (AA_3) {\scriptsize...};
\node [box] at (-1.6,-3) (NN_1) {\scriptsize...};
\node [box] at (-0.5,-3) (NN_2) {\scriptsize Ordering};
\node [box] at (0.6,-3) (NN_3) {\scriptsize...};
\node [box] at (1.7,-3) (GG_1) {\scriptsize...};
\node [box] at (3.3,-3) (GG_2) {\scriptsize Circle theorems};
\node [box] at (4.9,-3) (GG_3) {\scriptsize...};

\node [] at (6,0) (Level 0) {\scriptsize Level 0};
\node [] at (6,-1) (Level 1) {\scriptsize Level 1};
\node [] at (6,-2) (Level 2) {\scriptsize Level 2};
\node [] at (6,-3) (Level 3) {\scriptsize Level 3};

\path [arrowline] (Maths) to (Number);
\path [arrowline] (Maths) to (Algebra);
\path [arrowline] (Maths) to (Geometry);

\path [arrowline] (Algebra) to (A_1);
\path [arrowline] (Algebra) to (A_2);
\path [arrowline] (Algebra) to (A_3);
\path [arrowline] (Number) to (N_1);
\path [arrowline] (Number) to (N_2);
\path [arrowline] (Number) to (N_3);
\path [arrowline] (Geometry) to (G_1);
\path [arrowline] (Geometry) to (G_2);
\path [arrowline] (Geometry) to (G_3);

\path [arrowline] (A_2) to (AA_1);
\path [arrowline] (A_2) to (AA_2);
\path [arrowline] (A_2) to (AA_3);
\path [arrowline] (N_2) to (NN_1);
\path [arrowline] (N_2) to (NN_2);
\path [arrowline] (N_2) to (NN_3);
\path [arrowline] (G_2) to (GG_1);
\path [arrowline] (G_2) to (GG_2);
\path [arrowline] (G_2) to (GG_3);
\end{tikzpicture}
\vspace{-2mm}
    \caption{Hierarchy of topics in Eedi data. All the questions are related to maths (level 0 topic). The number of topics at levels 1, 2 and 3 are 3, 25 and 57. Each question is associated with only one topic at level 3 (thus, to only one topic at any higher level).}
    \label{fig:eedi_hierarchy}
    \vspace{-4mm}
\end{figure*}
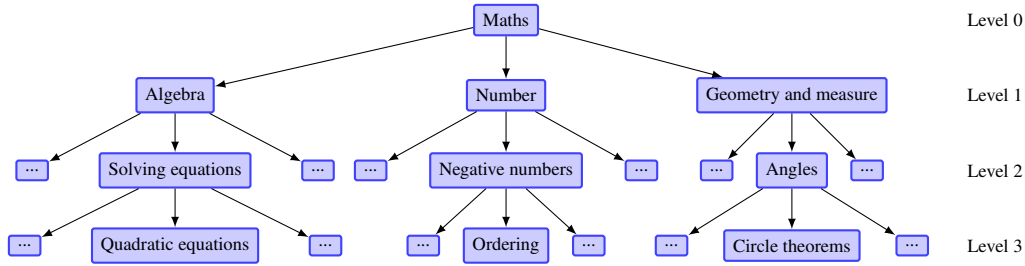

{Finally, we evaluate our method on an even more challenging real-world dataset in education requiring group-wise structure discovery. This is }
an important real-world problem in the field of AI-powered educational systems \citep{wang2021results, wang2020educational}.
{In this setting,} we are interested in relationships between {topics while the observations are question-answer pairs under these topics. }
The {dataset} is very sparse, with 74.1\% of the values missing.
The dataset contains the responses by 6147 students to 948 mathematics questions.
The 948 variables are binary (1 if the student provided the correct answer and 0 otherwise). 
These 948 questions target very specific mathematical concepts and are grouped within a meaningful hierarchy of \emph{topics}; see \autoref{fig:eedi_hierarchy}.
Here we apply {our proposed model to find the relationships among the topics using the third level of the topic hierarchy (\autoref{fig:eedi_hierarchy}), resulting in 57 group-level} nodes.

\begin{table}
    \centering
    \scriptsize
    \begin{tabular}{r@{\hskip 3mm}c@{\hskip 2mm}c@{\hskip 2mm}c}
\toprule
{} &               Accuracy &               AUROC & AUPR               \\
\midrule
Majority vote & 0.6260$\pm$0.0000 & 0.6208$\pm$0.0000 &  0.7465$\pm$0.0000 \\
Mean imputing & 0.6260$\pm$0.0000 & 0.6753$\pm$0.0000 &  0.6906$\pm$0.0000 \\
MICE & 0.6794$\pm$0.0005 & 0.7453$\pm$0.0007 &  0.7483$\pm$0.0010 \\
Missforest & 0.6849$\pm$0.0005 & 0.7219$\pm$0.0007 &  0.7478$\pm$0.0008 \\
PVAE & 0.7138$\pm$0.0005  & \textbf{0.7852$\pm$0.0001}  & \textbf{0.8204$\pm$0.0002} \\
\modelName &  \textbf{0.7147$\pm$0.0007} &  0.7815$\pm$0.0008 & 0.8179$\pm$0.0006 \\
\bottomrule
    \end{tabular}
    \vspace{-3mm}
    \caption{Imputation results for Eedi topics dataset (mean and standard error over five runs). \label{tab:eedi_imputation}}
\vspace{-5mm}
\end{table}

\begin{table}[h]
\centering
    \scriptsize
    \begin{tabular}{r@{\hskip 2mm}c@{\hskip 2mm}c@{\hskip 2mm}c@{\hskip 2mm}c}
    \toprule
    & \multicolumn{2}{c}{Adjacency} & \multicolumn{2}{c}{Orientation} \\
        \cmidrule[0.5pt](r){2-3}
        \cmidrule[0.5pt](l){4-5}
        & Expt 1 & Expt 2 & Expt 1 & Expt 2 \\
        \midrule
    \textit{Random}    & 2.04 & 2.08& 1.44  & 1.40 \\ 
    DAG-GNN & 2.04 & 2.32 & 1.68  & 1.68 \\
    \modelName    & \textbf{3.60} &  \textbf{3.70} & \textbf{2.76} & \textbf{2.60}  \\
    \bottomrule
        \end{tabular}
        \vspace{-3mm}
    \captionof{table}{Average expert evaluation of the topic relationships. Cohen's $\kappa$ inter-annotator agreement is $0.72$ for adjacency and $0.76$ for orientation (substantial agreement).\label{tab:relationships_summary}}
\end{table}
\begin{table*}[t]
    \scriptsize
    \centering
    \begin{tabular}{c@{\hskip 2mm}c@{\hskip 2mm}c}
    \begin{tabular}{r @{\hskip 1mm}c@{\hskip 1mm}c@{\hskip 1mm}c}
\toprule
\textit{\modelName} &  Number &  Algebra &  Geometry \\
\midrule
Number               &      30 &        4 &                     3 \\
Algebra              &       2 &        6 &                     0 \\
Geometry &       0 &        0 &                     5 \\
\bottomrule
\end{tabular}
        
         &
         
    \begin{tabular}{r @{\hskip 1mm}c@{\hskip 1mm}c@{\hskip 1mm}c}
\toprule
DAG-GNN &  Number &  Algebra &  Geometry \\
\midrule
Number               &       8 &        3 &                     6 \\
Algebra              &       1 &        5 &                     2 \\
Geometry &       14 &        7 &                    11 \\
\bottomrule
\end{tabular}   

&

    \begin{tabular}{r @{\hskip 1mm}c@{\hskip 1mm}c@{\hskip 1mm}c}
\toprule
\textit{Random} &  Number &  Algebra &  Geometry \\
\midrule
Number               &       7 &        4 &                     6 \\
Algebra              &       8 &        1 &                     6 \\
Geometry &       6 &        3 &                     9 \\
\bottomrule
\end{tabular}
    \end{tabular}
    \vspace{-3mm}
    \captionof{table}{Distribution of the relationships across level 1 topics.
    The item $(i,j)$ refers to edges in the direction $i\to j$.
    \label{tab:eedi_adj_level_1}
    }
    \vspace{-5mm}
\end{table*}

\textbf{Imputation results.}
\modelName achieves competitive or superior performance when compared to the baselines (\autoref{tab:eedi_imputation}).
Although the dataset is relatively balanced (54\% of the values are 1), we provide AUROC and AUPR for completeness.
Notice that this setting is more challenging than the previous ones since we learn relationships between groups of variables (topics). 
Indeed, whereas the group extension allows for more meaningful relationships, the information flow happens at a less granular level.
Interestingly, even in this case, \modelName obtains similar or improved imputation results compared to the baselines.

\textbf{Structure discovery results between groups.}
Most of the baselines used so far cannot be applied here because i) they cannot deal with partially observed training data or ii) they cannot learn relationships between groups of variables.
DAG-GNN is the only one that can be adapted to satisfy both properties. 
For the first one, we adapt DAG-GNN following the same strategy as in \modelName, i.e. replacing missing values with a constant value. 
For the second one, notice that DAG-GNN can be used for vector-valued variables according to the original formulation \citep{yu2019dag}. However, all of them need to have the same dimensionality. To cope with arbitrary groups, we apply the group-specific mappings described for \modelName. 
Finally, as an additional reference, we also compare with randomly generated relationships, which we refer to as \textit{Random}. 

Moreover, as there are no ground truth relationships in this real-world application,  we ask two experts (teachers)
to assess the validity of the relationships found by \modelName, DAG-GNN, and \textit{Random}.
For each relationship, they evaluate the adjacency (whether it is sensible to connect the two topics) and the orientation (whether the first one is a prerequisite for the second one).
They provide an integer value from 1 (strongly disagree) to 5 (strongly agree), i.e. the higher, the better.
The complete list of relationships and expert evaluations for \modelName, DAG-GNN, and \textit{Random} can be found in the appendix; see \autoref{tab:full_relationships_vicause}, \autoref{tab:full_relationships_dag_gnn}, and \autoref{tab:full_relationships_random}, respectively. 
In summary, \autoref{tab:relationships_summary} shows here the average evaluations: we see that the relationships discovered by \modelName score much more highly across both metrics than the baseline models. 

Another interesting aspect is how the relationships between level-3 topics are distributed across higher-level topics (recall \autoref{fig:eedi_hierarchy}).
Intuitively, it is expected that most of the relationships happen \emph{inside} higher-level topics (e.g. Number-related concepts are more probably related to each other than to Geometry-related ones). 
\autoref{tab:eedi_adj_level_1} shows such a distribution for the compared methods.
Indeed, notice that the percentage of inside-topic relationships is higher for \modelName (82\%) and DAG-GNN (42\%) than for \textit{Random} (34\%). 
An analogous analysis for the 25 level-2 topics is provided in the appendix; see \autoref{tab:eedi_adj_level_2_vicause} (\modelName), \autoref{tab:eedi_adj_level_2_dag_gnn} (DAG-GNN), and \autoref{tab:eedi_adj_level_2_random} (\textit{Random}).
In particular, whereas 6\% of the connections happen inside level 2 topics for \textit{Random}, it is 14\% for DAG-GNN and 36\% for \modelName.

{\textbf{Education Impact.}} Lastly, {to make a real-world impact, we have been provided with an additional education dataset in the same format as Eedi by an education organization to help provide insight for math curriculum building. The final structure among all topics found by \modelName is presented by figure \autoref{fig:eedi_topic_relationship} in the appendix. } 
The predicted relationships allowed insights into which topics are foundational and need to be covered earlier (topics with many originating edges), as well as which topics are more complex and should be covered later (topics with many incoming edges). This allowed us to re-evaluate the order of topics in a nationwide used secondary curriculum. Specifically, topics such as ``arithmetic'' or ``properties of shapes'' were moved earlier in the curriculum, while topics such as ``negative numbers'' or ``proportion and similarity'' were moved to a later stage in the curriculum. {Another interesting example found by the domain expert is the Venn diagram, which was originally taught in year 9/10 and is now suggested to move to year 7. Experts found that the Venn diagram has been a useful tool in teaching other topics which are currently taught before year 10.  Moving this earlier will help students to learn other topics better.} This emphasises the real-world impact our model, \modelName, can have in planning curricula.

%% file: conclusion.tex
\vspace{-3mm}
\section{Conclusions}\label{sec:conclusions}
\vspace{-2mm}
We introduced \modelName, a novel approach that simultaneously performs {group-wise} structure discovery and learns to impute missing values.
Both tasks are performed jointly: imputation is informed by the discovered relationships and vice-versa, {leading to improved performance for both tasks}.
Moreover, motivated by a real-world problem, \modelName { shows its impact in the real-world education domain to aid domain experts in setting up curriculum. }

%% file: appendix.tex
\section{\modelName for missing-at-random scenario}
\label{app: MAR}
In this section, we briefly explain why \modelName can handle MAR problem by leveraging the results from \citet{rubin1976inference}. Let's denote $\boldsymbol{r}_i$ as the missing mask, where $r_{i,d}=1$ indicates $x_{i,d}$ is observed. For the random variable $R$, we use $p_\lambda(\boldsymbol{r}|\bx)$ as the missing mechanism with parameters $\lambda$. To explicitly state the dependence of \modelName and its model parameter $\theta$, we use $p_\theta(\bx)$ to denote the corresponding model density. We can now formally define the concept of MAR.
\begin{definition}[Missing at Random \citep{rubin1976inference}]
The missing data are missing at random if for each value of $\lambda$, $p_\lambda(\boldsymbol{r}|\bx)$ takes teh same value for all $\bx_u$. Namely, $p_\lambda(\boldsymbol{r}|\bx)=p_\lambda(\boldsymbol{r}|\bx_o)$. 
\end{definition}

Recall that our \modelName is trained by maximizing the ELBO (Eq.\ref{eq:ELBO}) based on the observed values $x_o$. However, this formulation ignores the missing mechanisms $p_\phi(\boldsymbol{r}|\bx)$. In order to perform missing value imputation, one need to ensure that the inference for $\theta$ should be correct. In the following, we show under MAR, ignoring missing mechanism does not affect the correctness of inferring $\theta$ under ELBO. The following proof is an adaptation of Theorem 7.1 in \citet{rubin1976inference}. 

When explicitly modelling the missing mechanism, the joint likelihood can be written as 
\begin{align*}
    &\log p_{\theta,\phi}(\bx_o,\boldsymbol{r})\\
    &=\log \int p_\theta(\bx)p_\phi(\boldsymbol{r}|\bx)d\bx_u\\
    &=\log \int p_\theta(\bx,\boldsymbol{z},\boldsymbol{G})p_\phi(\boldsymbol{r}|\bx)d\boldsymbol{z}d\boldsymbol{G}d\bx_u\\
    &=\log p_\phi(\boldsymbol{r}|\bx_o)\int p_\theta(\bx,\boldsymbol{z},\boldsymbol{G})d\boldsymbol{z}d\boldsymbol{G}d\bx_u\\
    &\geq \log p_\phi(\boldsymbol{r}|\bx_o)+\text{ELBO}(\theta)
\end{align*}
where the third equality is from the definition of MAR and the last inequality is from the standard ELBO derivation. The above equation explicitly lower bounds the joint likelihood by two separate terms regarding $\theta$ and $\phi$. Thus, when performing inference over $\theta$, one can safely ignore the missing mechanism involving $\phi$, resulting in the same optimization objective as Eq.\ref{eq:ELBO}.

\section{Does \modelName respect the graph G in observational space}
\label{app: respect graph G}
From the formulation of the decoder in \modelName (Eq.\ref{eq:n2e} and \ref{eq:e2n}), the inferred graph $\bG$ seems to define whether the information flow between nodes is allowed or not. Namely, when $G_{ij}=1$, the information is allowed to pass from $z_{i}$ to $z_j$ at each iteration $t$. Thus, $\bG$ directly defines a structure for latent space $\bZ$, and indirectly defines a structure in observation $\bX$ through the GNN updates and the final read-out layer. A natural question to ask is whether the resulting observations $\bx$ from \modelName also respect the graph $\bG$. 
In the following, we show that when GNN is in equilibrium and the read-out layer is invertible without additional observational noise ($\sigma_x=0$), the \modelName is in fact a SEM for observation $\bx$, which respects the graph $\bG$. {In the following, for the clarity of notations, we consider the structure learning between variables. For group-wise relations, it is trivial to generalize, since the going from variable-wise to group-wise only changes the read-out layer, where we use $M$ different MLPs instead of one. }

First, let's clarify what do we mean by "respect a graph $\bG$". 
\begin{definition}[Respect a graph $\bG$]
For a given \modelName model $p(\bx,\bz;\bG)$ with a specific graph $\bG$, we say the model $p(\bx;\bG)=\int p(\bx,\bz;\bG)d\bz$ respects the graph $\bG$ if it can be factorized 
\[
p(\bx;\bG)=\prod_{d=1}^Dp(x_d|PA(d);\bG),
\]
where $PA(d)$ is a set of parents of node $i$ specified by graph $\bG$. 
\end{definition}

\subsection{GNN at steady state}
From the GNN message passing equations, we can re-organize Eq.\ref{eq:n2e} and \ref{eq:e2n} into one equation:
\begin{equation}
    z_i^t=F(PA(i)^{t-1},z_i^{t-1}),
    \label{eq: abstract GNN update equation}
\end{equation}
where $PA(i)^{t-1}$ is a set of parents' value for node $i$ at iteration $t-1$, $z_i^t$ is the value for node $i$ at iteration t and $F(\cdot)$ represents the GNN message passing updates. 

The above equation resembles a fixed-point iteration procedure for function $F$. Indeed, under the context of GNN, this has been considered as a standard procedure to search for the equilibrium state due to its exponential convergence (\citealt{dai2018learning}, Eq.1; \citealt{gu2020implicit}, Eq.2(b)). Thus, we assume that the GNN updates $F$ has unique equilibrium states given the initial conditions $AN(i)^0\cup z_i^0$ for each $i$, where $AN(i)^0$ represents the initial values of ancestors for node $i$. For a sufficient condition of its existence, one can refer to \citet[Theorem 4.1]{gu2020implicit}. We note that this is only a sufficient condition, meaning that the GNN without the conditions in \citet{gu2020implicit} can still have equilibrium states. Since discussing a necessary and sufficient conditions for the existence of the equilibrium state is out of the scope of this paper, we simply made an assumption that function $F$ has steady states. The reason we consider the initial ancestor values rather than just parent values is due to the message passing nature, where the value $PA(i)^t$ contains the information from the nodes that is at most $t$-hops away. 

Since graph $G$ represents a DAG, one can always find a permutation $\pi$ of the original index $i=1,\ldots,D$ based on the topological order. For concise notations, we assume the identity permutation. When the GNN is in equilibrium, we can rewrite Eq.\ref{eq: abstract GNN update equation} as
\begin{equation}
    z_i^\infty = F(PA^\infty(i),z_i^\infty),
    \label{eq: Equilibrium GNN}
\end{equation}
where the superscript $\infty$ represents that steady state of the node. From the assumption, since the steady state for each $z_i^\infty$ depends on the initial values $AN(i)^0\cup z_i^0$, it is trivial to see that the steady state $PA^\infty(i)$ depends on $AN^0(i)$. Therefore, the steady state $z_i^\infty$ is uniquely determined by $PA^\infty(i)$ and $z_i^0$. Namely,
\begin{equation}
    z_i^\infty = H_i(PA^\infty(i),z_i^0)
    \label{eq: Equilibrium GNN z_0}
\end{equation}
for $i=1,\ldots,D$, where $H_i$ is a mapping from $PA(i)^\infty$ and $z_i^0$ to the steady state of node $i$. 

This is exactly the general form of an \emph{structural equation model} (SEM) defined by graph $G$. If we further assume that the read-out layer $g(\cdot)$ is invertible, we can obtain
\begin{equation}
    x_i = g\left(H_i\left(g^{-1}\left(PA^\infty_x(i)\right),z_i^0\right)\right),
    \label{eq: Equilibrium GNN x}
\end{equation}
which is also an SEM based on $G$ for observation $\bx$. Thus, it is trivial that \modelName respects the graph $G$ based on the above assumptions. 

In practice, due to the exponential convergence of fixed point iteration, we found out that one does not need to use large iteration $t$. To balance the performance and computational cost, we found that $3$ iterations of GNN message passing is enough to obtain reasonable performances. 

\section{Experimental details}\label{app:details}

Here we specify the complete experimental details for full reproducibility. 
We first provide all the details for the synthetic experiment. Then we explain the differences for the neuropathic pain and the Eedi topics experiments.

\subsection{Synthetic experiment}\label{app:synthetic}

\textbf{Data generation process}.
To understand how the number of variables affects \modelName, we use $D=5,7,9$ variables (five datasets for each value of $D$).
We first sample the underlying true structure. 
An edge from variable $i$ to variable $j$ is sampled with probability $0.5$ if $i<j$, and probability $0$ if $i\ge j$ (this ensures that the true structure is a DAG, which is just a standard scenario, and not a requirement for any of the compared algorithms).
Then, we generate the data points.
Root nodes (i.e. nodes with no parents, like variables 1 and 2 in \autoref{fig:synthetic_graph}(a) in the paper) are sampled from $\mathcal{N}(0,1)$.
Any other node $v_i$ is obtained from its parents $\mathrm{Pa}(i)$ as $v_i=\sum_{j\in\mathrm{Pa}(i)}\sin(3v_j) + \varepsilon$, where $\varepsilon\rightarrow\mathcal{N}(0,0.01)$ is a Gaussian noise.
We use the $\sin$ function to induce non-linear relationships between variables. Notice that the $3$-times factor inside the $\sin$ encourages that the whole period of the $\sin$ function is used (to favor non-linearity).
To evaluate the imputation methods, 30\% of the test values are dropped.
As an example of the data generation process, \autoref{fig:synthetic_data} below shows the pair plot for the dataset generated from the graph in \autoref{fig:synthetic_graph}(a) in the paper. 

\textbf{Model parameters}.
We start by specifying the parameters associated to the generative process.
We use a prior probability $p_{ij}=0.05$ in $\p(\bG)$ for all the edges. This favours sparse graphs, and can be adjusted depending on the problem at hand.
The prior $\p(\bZ)$ is a standard Gaussian distribution, i.e. $\sigma_z^2=1$. This provides a standard regularisation for the latent space.
The output noise is set to $\sigma_x^2=0.02$, which favours the accurate reconstruction of samples.
As for the decoder, we perform $T=3$ iterations of GNN message passing. All the MLPs in the decoder (i.e. MLP$^f$, MLP$^b$, MLP$^{e2n}$ and $g$) have two linear layers with ReLU non-linearity.
The dimensionality of the hidden layer, which is the dimensionality of each latent subspace, is $256$.
Regarding the encoder, it is given by a multi-head neural network that defines the mean and standard deviation of the latent representation. The neural network is a MLP with two standard linear layers with ReLu non-linearity. The dimension of the hidden layer is also $256$.
When using groups, there are as many such MLPs as groups.
Finally, recall that the variational posterior $\q(\bG)$ is the product of independent Bernoulli distributions over the edges, with a probability $\bG_{ij}$ to be estimated for each edge. These values are all initialised to $\bG_{ij}=0.5$.

\textbf{Training hyperparameters}.
We use Adam optimizer with learning rate $0.001$. We train during $300$ epochs with a batch size of $100$ samples. Each one of the two stages described in the two-step training takes half of the epochs. The percentage of data dropped during training for each instance is sampled from a uniform distribution. 
When doing the reparametrization trick (i.e. when sampling from $\bZ_n$), we obtain one sample during training ($100$ samples in test time).
For the Gumbel-softmax sample, we use a temperature $\tau=0.5$. The rest of hyperparameters are the standard ones in \texttt{torch.nn.functional.gumbel\_softmax}, in particular we use soft samples.
To compute the DAG regulariser $\mathcal{R}(\bG)$, we use the exponential matrix implementation in \texttt{torch.matrix\_exp}.
This is in contrast to previous approaches, which resort to approximations \cite{zheng2018dags, yu2019dag}.
When applying the encoder, missing values in the training data are replaced with the value $0$ (continuous variables).

\textbf{Baselines details}.
Regarding the structure learning baselines, we ran both PC and GES with the Causal Command tool offered by the Center for Causal Discovery \url{https://www.ccd.pitt.edu/tools/}.
We used the default parameters in each case (i.e. disc-bic-score for GES and cg-lr-test for PC).
NOTEARS (L), NOTEARS (NL) and DAG-GNN were run with the code provided by the authors in GitHub: \url{https://github.com/xunzheng/notears} (NOTEARS (L) and NOTEARS (NL)) and \url{https://github.com/fishmoon1234/DAG-GNN} (DAG-GNN).
In all cases, we used the default parameters proposed by the authors.
Regarding the imputation baselines, Majority Vote and Mean Imputing were implemented in Python. 
MICE and Missforest were used from Scikit-learn library with default parameters \url{https://scikit-learn.org/stable/modules/generated/sklearn.impute.IterativeImputer.html#sklearn.impute.IterativeImputer}.
For PVAE, we use the authors implementation with their proposed parameters, see \url{https://github.com/microsoft/EDDI}. 

\textbf{Other experimental details}. \modelName is implemented in PyTorch. The code is available in the supplementary material. The experiments were run using a local Tesla K80 GPU and a compute cluster provided by Azure Machine Learning platform with NVIDIA Tesla V100 GPU.

\subsection{Neuropathic pain experiment}\label{app:neuropathic}

\textbf{Data generation process}.
We use the Neuropathic Pain Diagnosis Simulator in \url{https://github.com/TURuibo/Neuropathic-Pain-Diagnosis-Simulator}.
We simulate five datasets with 1500 samples, and split each one randomly in 1000 training and 500 test samples. To evaluate the imputation methods, 30\% of the test values are dropped. These five datasets are used for the five independent runs reported in experimental results.

\textbf{Model and training hyperparameters}.
Most of the hyperparameters are identical to the synthetic experiment. However, in this case we have to deal with 222 variables, many more than before.
In particular, the number of possible edges is 49062.
Therefore, we reduce the dimensionality of each latent subspace to $32$, the batch size to $25$, and the amount of test samples for $\bZ_n$ to $10$ (in training we still use one as before).  
Moreover, we reduce the initial posterior probability for each edge to $0.2$.
The reason is that, for $0.5$ initialization, the DAG regulariser $\mathcal{R}(\bG)$ evaluates to extremely high and unstable values for the $222\times 222$ matrix.
Since this is a more complex problem (no synthetic generation), we run the algorithm for $1000$ epochs.
When applying the encoder, missing values in the training data are replaced with the value $0.5$ (binary variables).

\subsection{Eedi topics experiment}\label{app:eedi}
\textbf{Data pre-processing}.

\textbf{Data generation process}.
The real-world Eedi topics dataset contains 6147 samples, and can be downloaded from the website \url{https://eedi.com/projects/neurips-education-challenge} (task3\_4 folder). The mapping from each question to its topics (also called "subjects") is given by the file ``data/metadata/question\_metadata\_task\_3\_4.csv''. For those questions that have more than one topic associated at the same level, randomly sample one of them.
The hierarcy of topics (recall \autoref{fig:eedi_hierarchy} in the paper) is given by the file ``data/metadata/subject\_metadata.csv''.
We use a random 80\%-10\%-10\% train-validation-test split. The validation set is used to perform Bayesian Optimization (BO) as described below.
The five runs reported in the experimental section come from different (random) initializations for the model parameters.

\textbf{Model and training hyperparameters}.
Here, we follow the same specifications as in the neuropathic pain dataset.
The only difference is that we perform BO for three hyperparameters: the dimensionality of the latent subspaces, the number of GNN message passing iterations, and the learning rate.
The possible choices for each hyperparameter are $\{5, 10, 15, 20, 25, 30, 35, 40, 45, 50\}$, $\{3, 5, 8, 10, 12, 14, 16, 18, 20\}$, and $\{10^{-4}, 10^{-3}, 10^{-2}\}$ respectively.
We perform $39$ runs of BO with the hyperdrive package in Azure Machine Learning platform \url{https://docs.microsoft.com/en-us/python/api/azureml-train-core/azureml.train.hyperdrive?view=azure-ml-py}.
We use validation accuracy as the target metric.
The best configuration obtained through BO was $15$, $8$ and $10^{-4}$, respectively.

\textbf{Baselines details}.
As explained in the paper, in this experiment DAG-GNN is adapted to deal with missing values and groups of arbitrary size. For the former, we adapt the DAG-GNN code to replace missing values with $0.5$ constant value, as in \modelName.
For the latter, we also follow \modelName and use as many different neural networks as groups (as described in the paper), all of them with the same architecture as the one used in the original code (\url{https://github.com/fishmoon1234/DAG-GNN}).

\textbf{Other experimental details}.
The list of relationships found by \modelName (\autoref{tab:full_relationships_vicause}) and DAG-GNN (\autoref{tab:full_relationships_dag_gnn}) aggregates the relationships obtained in the five independent runs.
This is done by setting a threshold of $0.35$ on the posterior probability of edge (which is initialized to $0.2$) and considering the union for the different runs.
This resulted in $50$ relationships for \modelName and $57$ for DAG-GNN. For \textit{Random}, we simulated $50$ random relationships.
Also, the probability reported in the first column of \autoref{tab:full_relationships_vicause} is the average of the probabilities obtained for that relationship in the five different runs.

\section{Additional figures and results}
\begin{figure}[h]
    \centering
    \includegraphics[width=0.5\textwidth]{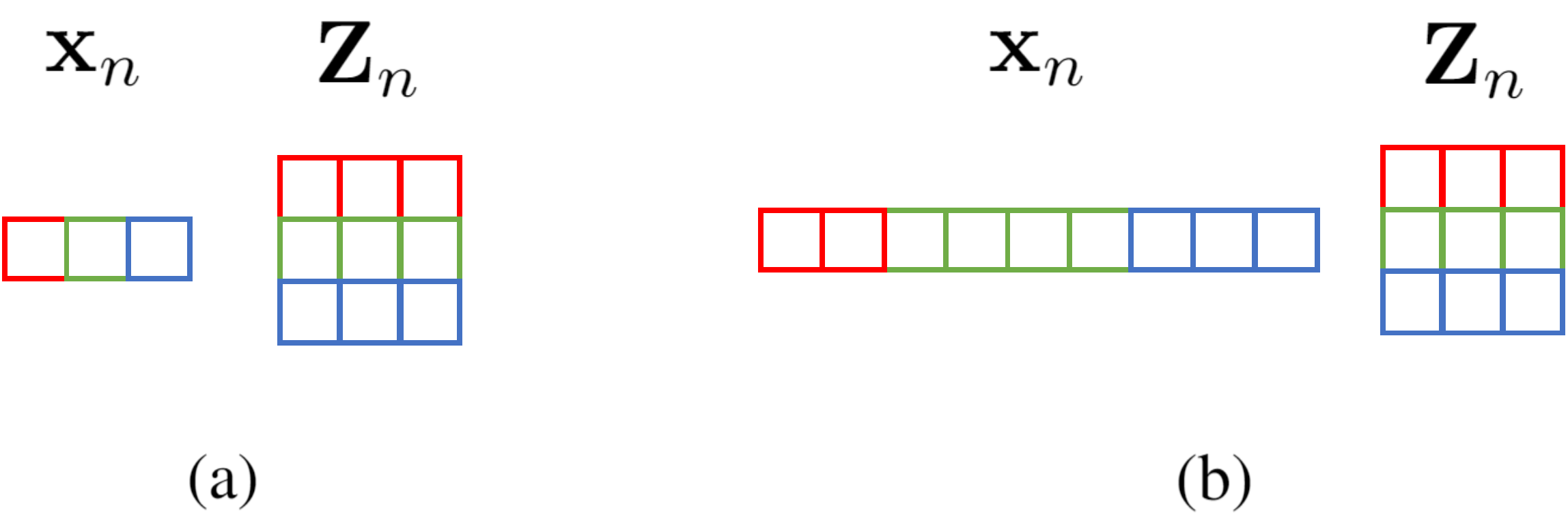}
    \vspace{3mm}
    \caption{Structured latent space. (a) At the level of variables. Each variable in $\bx_n$ (each color) has its own latent subspace, which is given by a row in $\bZ_n$. (b) At the level of groups of variables. Here, each group of variables (each color) has its own latent subspace, which is given by a row in $\bZ_n$.}
    \label{fig:structured_latent_space}
\end{figure}

\begin{figure}[h]
    \centering
    \includegraphics[width=0.5\textwidth]{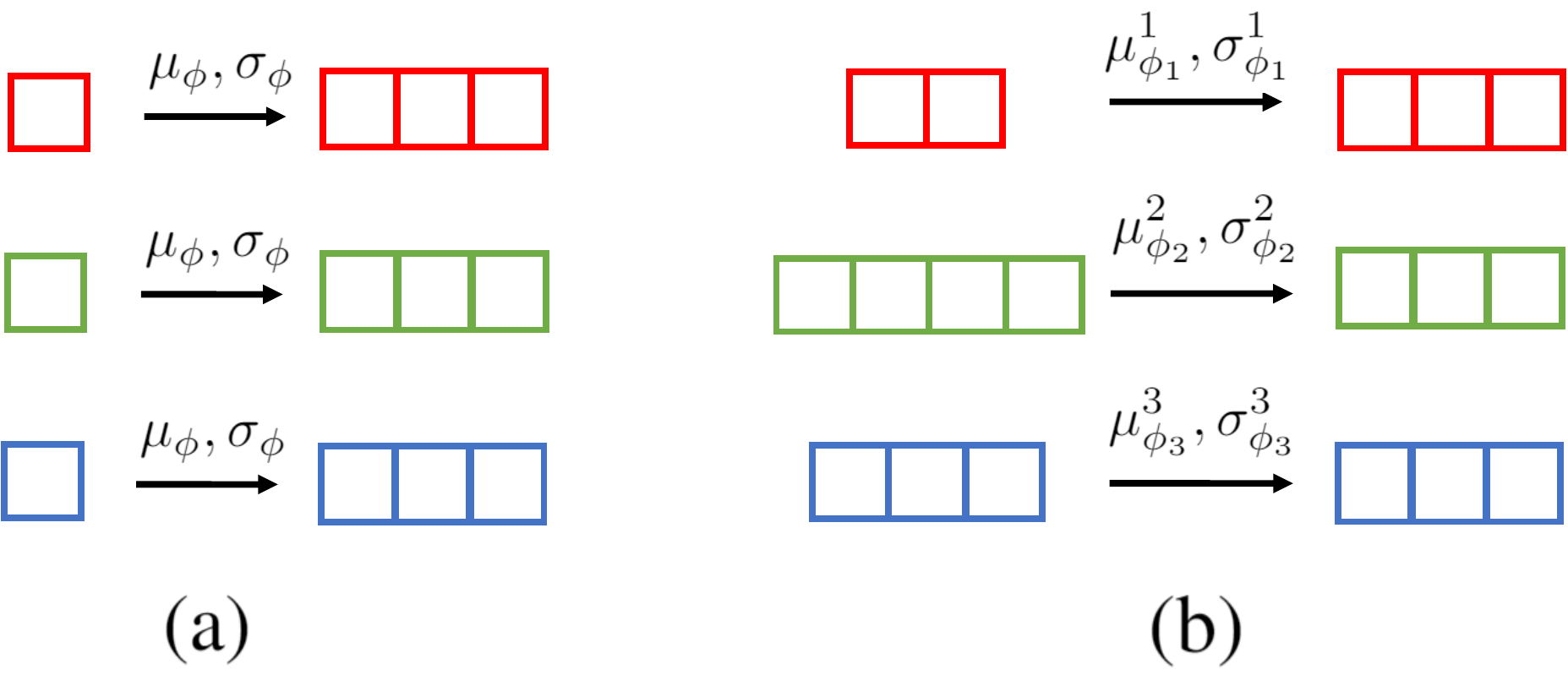}
    \caption{The encoder respects the structure of the latent space. (a) At the level of variables. All the variables use the same encoding functions. (b) At the level of groups of variables. Each group of variables uses different encoding functions.
    \label{fig:structured_mappings}}
\end{figure}

\begin{figure*}[h]
    \centering
    \includegraphics[width=0.6\textwidth]{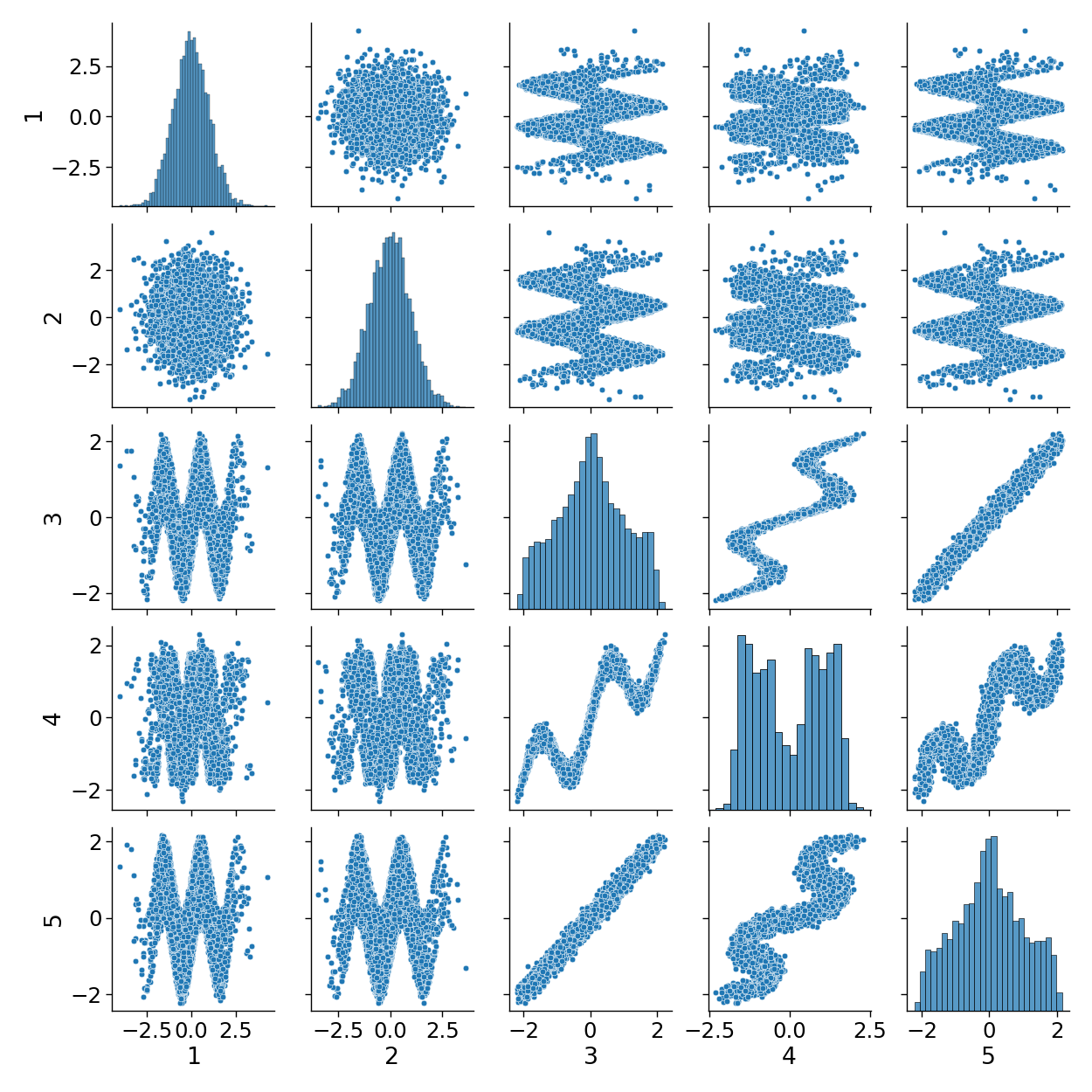}
    \caption{Pair-plot for the dataset generated from the graph in \autoref{fig:synthetic_graph}(a) in the paper. We observe different type of relationships between variables, including non-linear ones.}
    \label{fig:synthetic_data}
\end{figure*}


\begin{table*}[h]
    \centering
\begin{tabular}{rcccc}
\toprule
{} & \multicolumn{3}{c}{Number of variables} & \multirow{2}{*}{Average} \\
\cmidrule[0.5pt](lr){2-4}
{} &            5 &             7 &              9 & {}\\
\midrule
Majority vote &  0.5507$\pm$0.0056 &  0.5391$\pm$0.0050 &  0.5427$\pm$0.0050 &  0.5442$\pm$0.0032 \\
Mean imputing &  0.2351$\pm$0.0104 &  0.2124$\pm$0.0112 &  0.2143$\pm$0.0064 &  0.2206$\pm$0.0061 \\
MICE          &  0.1352$\pm$0.0044 &  0.1501$\pm$0.0095 &  0.1230$\pm$0.0025 &  0.1361$\pm$0.0046 \\
Missforest    &  0.1279$\pm$0.0040 &  0.1403$\pm$0.0030 &  0.1258$\pm$0.0022 &  0.1313$\pm$0.0025 \\
PVAE &  0.1324$\pm$0.0048 &  0.1536$\pm$0.0095 &   0.1360$\pm$0.0019 & 0.1407$\pm$0.0043 \\
\modelName &  \textbf{0.1146$\pm$0.0026} &  \textbf{0.1251$\pm$0.0055} &  \textbf{0.1191$\pm$0.0015} & \textbf{0.1196$\pm$0.0024} \\
\bottomrule
\end{tabular}
    \caption{Imputation results for the synthetic experiment in terms of RMSE (not aggregating by number of variables, $D=5,7,9$). The values are the mean and standard error over five different simulations.
    }
    \label{tab:synthetic_imputation_extended}
\end{table*}

\begin{table*}[h]
    \centering
    \begin{tabular}{rl}
\toprule
 Index &                                     Topic name \\
\midrule
     1 &                                       Decimals \\
     2 &                  Factors, Multiples and Primes \\
     3 & Fractions, Decimals and Percentage Equivalence \\
     4 &                                      Fractions \\
     5 &                      Indices, Powers and Roots \\
     6 &                               Negative Numbers \\
     7 &                           Straight Line Graphs \\
     8 &                                   Inequalities \\
     9 &                                      Sequences \\
    10 &            Writing and Simplifying Expressions \\
    11 &                                         Angles \\
    12 &                                        Circles \\
    13 &                                   Co-ordinates \\
    14 &           Construction, Loci and Scale Drawing \\
    15 &                                       Symmetry \\
    16 &                           Units of Measurement \\
    17 &                        Volume and Surface Area \\
    18 &                               Basic Arithmetic \\
    19 &                                    Factorising \\
    20 &                              Solving Equations \\
    21 &                                        Formula \\
    22 &              2D Names and Properties of Shapes \\
    23 &                             Perimeter and Area \\
    24 &                      Similarity and Congruency \\
    25 &                                Transformations \\
\bottomrule
\end{tabular}

    \caption{Mapping between indexes for row/column names in \autoref{tab:eedi_adj_level_2_vicause} and \autoref{tab:eedi_adj_level_2_random} and the actual level-2 topic names.}
    \label{tab:map_level_2}
\end{table*}

\begin{table*}
    \centering
    \small
    \setlength{\tabcolsep}{3pt}
    \begin{tabular}{crccccccc}
\toprule
{} & {} &  \multicolumn{3}{c}{Adjacency} & \multicolumn{3}{c}{Orientation} & \multirow{2}{*}[-0.2em]{\shortstack{Causal\\Accuracy}} \\ 
\cmidrule[0.5pt](lr){3-5}
\cmidrule[0.5pt](lr){6-8}
{} & {} &               Recall &              Precision &               F$_{1}$-score &               Recall &  Precision &               F$_{1}$-score &               {} \\
\midrule
\multirow{6}{*}{5} & PC &  0.464$\pm$0.099 &   0.610$\pm$0.117 &  0.526$\pm$0.107 &  0.364$\pm$0.098 &   0.490$\pm$0.127 &  0.416$\pm$0.111 &  0.436$\pm$0.076 \\
{} & GES &  0.414$\pm$0.067 &  0.507$\pm$0.071 &  0.446$\pm$0.065 &  0.257$\pm$0.103 &  0.327$\pm$0.117 &   0.285$\pm$0.110 &  0.368$\pm$0.072 \\
{} & NOTEARS (L) & 0.186$\pm$0.052 &  0.400$\pm$0.089 &  0.247$\pm$0.063 &  0.119$\pm$0.049 &  0.300$\pm$0.110 &  0.167$\pm$0.065 &  0.119$\pm$0.049 \\
{} & NOTEARS (NL) & 0.331$\pm$0.057 &  0.470$\pm$0.078 &  0.384$\pm$0.065 &  0.264$\pm$0.047 &  0.370$\pm$0.053 &  0.304$\pm$0.049 &  0.264$\pm$0.047 \\
{} & DAG-GNN &  0.381$\pm$0.130 &  0.433$\pm$0.121 &  0.399$\pm$0.127 &  0.231$\pm$0.067 &  0.283$\pm$0.073 &  0.249$\pm$0.068 &  0.231$\pm$0.067 \\
{} & \modelName &  0.971$\pm$0.026 &  0.598$\pm$0.059 &   0.730$\pm$0.047 &  0.574$\pm$0.111 &  0.356$\pm$0.085 &  0.432$\pm$0.093 &  0.971$\pm$0.026 \\
\midrule
\multirow{6}{*}{7} & PC &   0.396$\pm$0.110 &  0.639$\pm$0.154 &  0.468$\pm$0.112 &  0.113$\pm$0.043 &  0.193$\pm$0.083 &   0.134$\pm$0.050 &  0.324$\pm$0.088 \\
{} & GES &  0.429$\pm$0.087 &  0.647$\pm$0.042 &  0.501$\pm$0.076 &  0.208$\pm$0.067 &  0.279$\pm$0.081 &  0.235$\pm$0.073 &  0.345$\pm$0.091 \\
{} & NOTEARS (L) & 0.222$\pm$0.059 &  0.526$\pm$0.124 &  0.309$\pm$0.078 &  0.176$\pm$0.041 &  0.436$\pm$0.109 &  0.248$\pm$0.058 &  0.176$\pm$0.041 \\
{} & NOTEARS (NL) & 0.315$\pm$0.094 &  0.513$\pm$0.119 &  0.382$\pm$0.104 &  0.269$\pm$0.074 &  0.453$\pm$0.105 &  0.330$\pm$0.084 &  0.269$\pm$0.074 \\ 
{} & DAG-GNN &  0.396$\pm$0.109 &  0.539$\pm$0.123 &  0.446$\pm$0.111 &  0.318$\pm$0.082 &  0.445$\pm$0.102 &  0.361$\pm$0.085 &  0.318$\pm$0.082 \\
{} & \modelName &  0.813$\pm$0.088 &  0.694$\pm$0.057 &  0.725$\pm$0.053 &  0.559$\pm$0.134 &   0.447$\pm$0.070 &   0.480$\pm$0.089 &  0.701$\pm$0.103 \\
\midrule
\multirow{6}{*}{9} & PC &  0.406$\pm$0.072 &  0.654$\pm$0.053 &   0.491$\pm$0.060 &   0.176$\pm$0.020 &  0.302$\pm$0.045 &  0.219$\pm$0.024 &  0.229$\pm$0.041 \\
{} & GES &  0.514$\pm$0.065 &   0.553$\pm$0.050 &  0.525$\pm$0.049 &  0.282$\pm$0.057 &  0.308$\pm$0.068 &  0.291$\pm$0.061 &  0.379$\pm$0.069 \\
{} & NOTEARS (L) & 0.172$\pm$0.026 &  0.403$\pm$0.076 &  0.238$\pm$0.036 &  0.151$\pm$0.023 &  0.366$\pm$0.082 &  0.211$\pm$0.035 &  0.151$\pm$0.023 \\
{} & NOTEARS (NL) & 0.338$\pm$0.042 &  0.485$\pm$0.053 &  0.394$\pm$0.045 &  0.297$\pm$0.034 &  0.429$\pm$0.044 &  0.347$\pm$0.036 &  0.297$\pm$0.034 \\
{} & DAG-GNN &  0.551$\pm$0.067 &  0.554$\pm$0.053 &  0.547$\pm$0.057 &  0.508$\pm$0.061 &  0.516$\pm$0.054 &  0.508$\pm$0.055 &  0.508$\pm$0.061 \\
{} & \modelName &  0.705$\pm$0.061 &  0.615$\pm$0.042 &  0.652$\pm$0.044 &  0.356$\pm$0.092 &  0.297$\pm$0.065 &  0.322$\pm$0.076 &  0.526$\pm$0.081 \\
\bottomrule
\end{tabular}
    \caption{Structure learning results for the synthetic experiment (not aggregating by number of variables, $D=5,7,9$). The values are the mean and standard error over five different simulations.}
    \label{tab:synthetic_causality_extended}
\end{table*}

\begin{table*}
\setlength{\tabcolsep}{3pt}
    \centering
    \small
\begin{tabular}{rllllllll}
\toprule
{} &  \multicolumn{8}{c}{Number of variables} \\
\cmidrule[0.5pt](lr){2-9}
{} &    4  & 8 & 16 & 32 & 64 & 128 & 256 & 512 \\
\midrule
PC           & 2.49$\pm$0.62  & 5.19$\pm$1.01  & 8.14$\pm$1.64   & 14.99$\pm$1.59   & 21.65$\pm$2.19  & 26.11$\pm$1.70  & 30.21$\pm$2.01  & 35.43$\pm$1.56  \\
GES          & 0.21$\pm$0.02  & 1.12$\pm$0.41  & 1.80$\pm$0.80   & 2.28$\pm$0.78    & 2.76$\pm$1.01   & 3.34$\pm$0.52   & 3.87$\pm$0.66   & 4.10$\pm$0.71   \\
NOTEARS (L)  & 8.91$\pm$2.34  & 21.04$\pm$3.43 & 38.53$\pm$2.52  & 56.11$\pm$3.23   & 91.11$\pm$4.15  & 140.33$\pm$3.53 & 331.04$\pm$6.55 & 378.21$\pm$9.12 \\
NOTEARS (NL) & 12.94$\pm$2.18 & 31.03$\pm$3.11 & 54.08$\pm$4.10  & 89.35$\pm$4.11   & 99.32$\pm$5.12  & 240.43$\pm$5.39 & 364.92$\pm$3.22 & 469.43$\pm$4.77 \\
DAG-GNN      & 13.62$\pm$2.93 & 30.04$\pm$2.48 & 52.01$\pm$3.81  & 88.12$\pm$79      & 112.33$\pm$5.01 & 255.11$\pm$6.93 & 371.22$\pm$5.32 & 498.09$\pm$5.01 \\
\modelName      & 10.27$\pm$1.98 & 25.11$\pm$5.21 & 47.98$\pm$3.12  & 76.12$\pm$4.40   & 101.12$\pm$4.23 & 201.59$\pm$6.33 & 340.10$\pm$8.22 & 421.11$\pm$5.33\\
MMHC         & 7.85$\pm$1.02  & 69.10$\pm$5.32 & 542.92$\pm$9.82 & 1314.76$\pm$9.10 & NA            & NA            & NA            & NA            \\
Tabu         & 2.01$\pm$0.72  & 7.45$\pm$1.03  & 24.08$\pm$5.93  & 57.87$\pm$3.85   & 77.87$\pm$5.52  & 128.67$\pm$4.09 & 163.33$\pm$6.55 & 219.05$\pm$3.42 \\
HillClimb         & 1.52$\pm$0.64  & 6.98$\pm$1.23  & 22.10$\pm$5.32  & 51.78$\pm$4.06   & 75.29$\pm$5.84  & 121.92$\pm$4.71 & 157.82$\pm$6.87 & 209.54$\pm$5.01 \\
\bottomrule
\end{tabular}
    \caption{Running times (in minutes) for different structure learning approaches in an extended synthetic experiment. For each number of variables, three datasets were simulated following the same data generation process described above, and the results show the mean and standard error. Notice that we have considered three additional baselines (MMHC, Tabu, HillClimb). We observe three different types of methods. MMHC scales poorly (NA means that the training took more than 24 hours), probably due to its hybrid nature that combines constraint-based and score-based approaches. \modelName and the other deep learning based methods (DAG-GNN, NOTEARS) can scale to large numbers of variables. Of course, simpler methods such as PC, GES, Tabu, and HillClimb are significantly faster than \modelName (note also that these baselines are from highly optimized libraries that leverage e.g. dynamic programming and parallelization), but their performance is worse. Indeed, the structure learning performance for this experiment is shown in \autoref{tab:causality_appendix}.\label{tab:synthetic_times}}
\end{table*}

\begin{table*}
    \centering
    \small
\begin{tabular}{rrrrrrrrrrrrrrrrrrrrrrrrrr}
\toprule
{} &  1  &  2  &  3  &  4  &  5  &  6  &  7  &  8  &  9  &  10 &  11 &  12 &  13 &  14 &  15 &  16 &  17 &  18 &  19 &  20 &  21 &  22 &  23 &  24 &  25 \\
\midrule
1  &   0 &   0 &   0 &   0 &   0 &   0 &   0 &   0 &   0 &   0 &   0 &   0 &   0 &   0 &   0 &   0 &   0 &   0 &   0 &   0 &   0 &   0 &   0 &   0 &   0 \\
2  &   0 &   2 &   0 &   0 &   1 &   0 &   0 &   0 &   0 &   0 &   0 &   0 &   0 &   0 &   0 &   0 &   0 &   4 &   0 &   0 &   0 &   0 &   0 &   0 &   0 \\
3  &   0 &   0 &   0 &   0 &   0 &   0 &   0 &   0 &   0 &   0 &   0 &   0 &   0 &   0 &   0 &   0 &   0 &   0 &   0 &   0 &   0 &   0 &   0 &   0 &   0 \\
4  &   0 &   0 &   0 &   0 &   0 &   0 &   0 &   0 &   0 &   0 &   0 &   0 &   0 &   0 &   0 &   0 &   0 &   0 &   0 &   0 &   0 &   0 &   0 &   0 &   0 \\
5  &   0 &   0 &   0 &   0 &   0 &   0 &   0 &   0 &   0 &   0 &   0 &   0 &   0 &   0 &   0 &   0 &   0 &   0 &   0 &   0 &   0 &   0 &   0 &   0 &   0 \\
6  &   0 &   5 &   0 &   0 &   1 &   6 &   0 &   0 &   0 &   2 &   1 &   0 &   0 &   0 &   0 &   1 &   0 &   4 &   0 &   0 &   2 &   0 &   0 &   0 &   0 \\
7  &   0 &   0 &   0 &   0 &   0 &   0 &   0 &   0 &   0 &   0 &   0 &   0 &   0 &   0 &   0 &   0 &   0 &   0 &   0 &   0 &   0 &   0 &   0 &   0 &   0 \\
\bottomrule
\end{tabular}
    \caption{How the 50 relationships found by \modelName are distributed across level 2 topics. The item $(i,j)$ refers to edges in the direction $i\to j$. There are 18 relationships inside level 2 topics (36\%). See \autoref{tab:map_level_2} for a mapping between indexes shown here in row/column names and the actual level-2 topic names.}
    \label{tab:eedi_adj_level_2_vicause}
\end{table*}

\begin{table*}
    \centering
    \small
\begin{tabular}{rrrrrrrrrrrrrrrrrrrrrrrrrr}
\toprule
{} &  1  &  2  &  3  &  4  &  5  &  6  &  7  &  8  &  9  &  10 &  11 &  12 &  13 &  14 &  15 &  16 &  17 &  18 &  19 &  20 &  21 &  22 &  23 &  24 &  25 \\
\midrule
1  &   0 &   1 &   0 &   0 &   0 &   0 &   0 &   0 &   0 &   0 &   0 &   0 &   0 &   0 &   0 &   0 &   0 &   0 &   0 &   0 &   0 &   0 &   0 &   0 &   0 \\
2  &   0 &   0 &   0 &   0 &   0 &   0 &   1 &   0 &   0 &   0 &   0 &   0 &   0 &   0 &   0 &   0 &   0 &   0 &   1 &   0 &   0 &   0 &   0 &   0 &   0 \\
3  &   0 &   0 &   0 &   0 &   0 &   1 &   0 &   0 &   0 &   0 &   0 &   0 &   0 &   0 &   0 &   0 &   1 &   1 &   0 &   0 &   0 &   0 &   0 &   0 &   0 \\
4  &   0 &   0 &   0 &   1 &   0 &   0 &   0 &   0 &   0 &   0 &   0 &   0 &   0 &   0 &   0 &   0 &   0 &   0 &   0 &   0 &   0 &   0 &   0 &   0 &   0 \\
5  &   0 &   0 &   0 &   0 &   0 &   1 &   0 &   1 &   0 &   0 &   0 &   0 &   0 &   0 &   0 &   0 &   1 &   0 &   0 &   0 &   0 &   0 &   0 &   0 &   0 \\
6  &   0 &   0 &   0 &   0 &   0 &   2 &   0 &   0 &   0 &   0 &   0 &   0 &   0 &   0 &   0 &   0 &   0 &   0 &   0 &   0 &   0 &   1 &   0 &   0 &   0 \\
7  &   0 &   0 &   0 &   0 &   0 &   0 &   0 &   0 &   0 &   0 &   0 &   0 &   0 &   0 &   0 &   0 &   0 &   0 &   0 &   0 &   0 &   0 &   0 &   0 &   0 \\
\bottomrule
\end{tabular}
    \caption{How the 57 relationships found by DAG-GNN are distributed across level 2 topics. The item $(i,j)$ refers to edges in the direction $i\to j$. There are 8 relationships inside level 2 topics (14\%). See \autoref{tab:map_level_2} for a mapping between indexes shown here in row/column names and the actual level-2 topic names.}
    \label{tab:eedi_adj_level_2_dag_gnn}
\end{table*}

\begin{table*}
    \centering
    \small
\begin{tabular}{rrrrrrrrrrrrrrrrrrrrrrrrrr}
\toprule
{} &  1  &  2  &  3  &  4  &  5  &  6  &  7  &  8  &  9  &  10 &  11 &  12 &  13 &  14 &  15 &  16 &  17 &  18 &  19 &  20 &  21 &  22 &  23 &  24 &  25 \\
\midrule
1  &   0 &   0 &   0 &   0 &   0 &   0 &   0 &   0 &   0 &   1 &   0 &   0 &   0 &   0 &   0 &   0 &   0 &   0 &   0 &   0 &   0 &   0 &   0 &   0 &   0 \\
2  &   0 &   0 &   0 &   0 &   0 &   0 &   0 &   0 &   0 &   0 &   0 &   0 &   0 &   0 &   0 &   0 &   0 &   0 &   0 &   0 &   0 &   0 &   0 &   0 &   0 \\
3  &   0 &   0 &   0 &   0 &   0 &   0 &   0 &   0 &   0 &   0 &   0 &   0 &   0 &   0 &   0 &   0 &   0 &   0 &   0 &   0 &   0 &   0 &   1 &   0 &   0 \\
4  &   0 &   3 &   0 &   1 &   0 &   0 &   1 &   0 &   0 &   0 &   0 &   0 &   0 &   0 &   0 &   0 &   1 &   0 &   0 &   0 &   0 &   0 &   0 &   0 &   0 \\
5  &   0 &   0 &   0 &   0 &   0 &   0 &   0 &   0 &   0 &   0 &   0 &   0 &   1 &   0 &   0 &   0 &   0 &   1 &   0 &   0 &   0 &   0 &   0 &   0 &   0 \\
6  &   0 &   0 &   0 &   0 &   0 &   0 &   0 &   0 &   0 &   0 &   0 &   1 &   1 &   0 &   0 &   0 &   0 &   0 &   0 &   0 &   0 &   1 &   0 &   0 &   0 \\
7  &   0 &   0 &   0 &   0 &   0 &   1 &   0 &   0 &   0 &   0 &   0 &   0 &   0 &   0 &   0 &   0 &   0 &   0 &   0 &   0 &   0 &   0 &   0 &   0 &   0 \\
\bottomrule
\end{tabular}
    \caption{How the 50 relationships found by \textit{Random} are distributed across level 2 topics. The item $(i,j)$ refers to edges in the direction $i\to j$. There are 3 relationships inside level 2 topics (6\%). See \autoref{tab:map_level_2} for a mapping between indexes shown here in row/column names and the actual level-2 topic names.}
    \label{tab:eedi_adj_level_2_random}
\end{table*}

\begin{landscape}

\newpage
\begin{table}
    \centering
    \tiny
    \begin{tabular}{rllrrrr}
\toprule
 Prob &                                                                                          Topic 1 (from) &                                                                                          Topic 2 (to) &  Adj1 &  Ori1 &  Adj2 &  Ori2 \\
\midrule
 0.44 &                              Adding and Subtracting Negative Numbers [Negative Numbers] [Number] &                                            Ordering Negative Numbers [Negative Numbers] [Number] &     5 &     1 &     5 &     1 \\
 0.38 &                                   Mental Multiplication and Division [Basic Arithmetic] [Number] &                    Multiples and Lowest Common Multiple [Factors, Multiples and Primes] [Number] &     5 &     5 &     5 &     5 \\
 0.37 &                                   Mental Multiplication and Division [Basic Arithmetic] [Number] &                       Factors and Highest Common Factor [Factors, Multiples and Primes] [Number] &     5 &     5 &     5 &     5 \\
 0.37 &                              Adding and Subtracting Negative Numbers [Negative Numbers] [Number] &                    Multiples and Lowest Common Multiple [Factors, Multiples and Primes] [Number] &     2 &     2 &     2 &     1 \\
 0.36 &                              Adding and Subtracting Negative Numbers [Negative Numbers] [Number] &                       Factors and Highest Common Factor [Factors, Multiples and Primes] [Number] &     2 &     1 &     2 &     1 \\
 0.35 &                                   Mental Multiplication and Division [Basic Arithmetic] [Number] &                                                          Place Value [Basic Arithmetic] [Number] &     4 &     2 &     4 &     2 \\
 0.35 &                                   Mental Multiplication and Division [Basic Arithmetic] [Number] &                                                               BIDMAS [Basic Arithmetic] [Number] &     5 &     5 &     5 &     5 \\
\bottomrule
\end{tabular}
    \caption{Full list of relationships found by \modelName in the Eedi topics dataset. Each row refers to one relationship (one edge). From left to right, the columns are the posterior probability of the edge, the sending node (topic), the receiving node (topic), and the adjacency and orientation evaluations from each expert. For each topic, the brackets contain its parent level 2 and level 1 topics.}
    \label{tab:full_relationships_vicause}
\end{table}
\end{landscape}

\begin{landscape}
\newpage
\begin{table}
    \centering
    \tiny
\begin{tabular}{llrrrr}
\toprule
                                                                                        Topic 1 (From) &                                                                                     Topic 2 (To) &  Adj1 &  Ori1 &  Adj2 &  Ori2 \\
\midrule
                                           Missing Lengths [Perimeter and Area] [Geometry and Measure] &                                       Midpoint Between Two Co-ordinates [Co-ordinates] [Algebra] &     4 &     4 &     4 &     5 \\
                      Construct Triangle [Construction, Loci and Scale Drawing] [Geometry and Measure] &                                                          Place Value [Basic Arithmetic] [Number] &     1 &     1 &     1 &     1 \\
                                              Squares, Cubes, etc [Indices, Powers and Roots] [Number] &                                Volume of Prisms [Volume and Surface Area] [Geometry and Measure] &     4 &     5 &     5 &     4 \\
Converting between Fractions and Percentages [Fractions, Decimals and Percentage Equivalence] [Number] &                                Volume of Prisms [Volume and Surface Area] [Geometry and Measure] &     1 &     1 &     1 &     1 \\
                                                   Angles in Triangles [Angles] [Geometry and Measure] &                                               Parts of a Circle [Circles] [Geometry and Measure] &     1 &     1 &     1 &     1 \\
                                          Types, Naming and Estimating [Angles] [Geometry and Measure] &                                  Angle Facts with Parallel Lines [Angles] [Geometry and Measure] &     4 &     5 &     5 &     5 \\
                                        Mental Multiplication and Division [Basic Arithmetic] [Number] &                                                 Measuring Angles [Angles] [Geometry and Measure] &     1 &     1 &     1 &     1 \\
\bottomrule
\end{tabular}
    \caption{Full list of relationships found by DAG-GNN in the Eedi topics dataset. Each row refers to one relationship (one edge). From left to right, the columns are the sending node (topic), the receiving node (topic), and the adjacency and orientation evaluations from each expert. For each topic, the brackets contain its parent level 2 and level 1 topics.}
    \label{tab:full_relationships_dag_gnn}
\end{table}
\end{landscape}

\begin{landscape}
\newpage
\begin{table}
    \centering
    \tiny
\begin{tabular}{llrrrr}
\toprule
                                                                                        Topic 1 (From) &                                                                                     Topic 2 (To) &  Adj1 &  Ori1 &  Adj2 &  Ori2 \\
\midrule
                                            Midpoint Between Two Co-ordinates [Co-ordinates] [Algebra] &                                              Angles in Triangles [Angles] [Geometry and Measure] &     1 &     1 &     1 &     1 \\
                                                  Solving Linear Inequalities [Inequalities] [Algebra] &                                             Enlargement [Transformations] [Geometry and Measure] &     1 &     1 &     1 &     1 \\
                                              Squares, Cubes, etc [Indices, Powers and Roots] [Number] &                                               Written Multiplication [Basic Arithmetic] [Number] &     4 &     1 &     5 &     1 \\
                                                         Substitution into Formula [Formula] [Algebra] &                                               Written Multiplication [Basic Arithmetic] [Number] &     4 &     1 &     3 &     1 \\
                                                     Linear Sequences (nth term) [Sequences] [Algebra] &                                   Mental Multiplication and Division [Basic Arithmetic] [Number] &     5 &     1 &     5 &     2 \\
                                                      Measuring Angles [Angles] [Geometry and Measure] &                    Construct Angle [Construction, Loci and Scale Drawing] [Geometry and Measure] &     5 &     5 &     5 &     5 \\
                                                               Dividing Fractions [Fractions] [Number] &                                Volume of Prisms [Volume and Surface Area] [Geometry and Measure] &     2 &     2 &     2 &     2 \\
\bottomrule
\end{tabular}
    \caption{Full list of relationships found by \textit{Random} in the Eedi topics dataset. Each row refers to one relationship (one edge). From left to right, the columns are the sending node (topic), the receiving node (topic), and the adjacency and orientation evaluations from each expert. For each topic, the brackets contain its parent level 2 and level 1 topics.}
    \label{tab:full_relationships_random}
\end{table}

\end{landscape}

\begin{table*}
\setlength{\tabcolsep}{4pt}
    \centering
    \small
\begin{tabular}{rccccccc}
\toprule
{} &  \multicolumn{3}{c}{Adjacency} & \multicolumn{3}{c}{Orientation} & \multirow{2}{*}[-0.2em]{\shortstack{Causal\\accuracy}} \\ 
\cmidrule[0.5pt](lr){2-4}
\cmidrule[0.5pt](lr){5-7}
{} &               Recall &              Precision &               F$_{1}$-score &               Recall &  Precision &               F$_{1}$-score &               {} \\
\midrule
PC           & 0.312$\pm$0.072 & 0.508$\pm$0.077 & 0.381$\pm$0.057 & 0.123$\pm$0.042 & 0.202$\pm$0.069 & 0.136$\pm$0.063 & 0.211$\pm$0.065 \\ 
GES          & 0.295$\pm$0.057 & 0.473$\pm$0.043 & 0.378$\pm$0.051 & 0.132$\pm$0.055 & 0.200$\pm$0.071 & 0.138$\pm$0.058 & 0.238$\pm$0.048 \\ 
NOTEARS (L)  & 0.123$\pm$0.047 & 0.401$\pm$0.063 & 0.219$\pm$0.040 & 0.093$\pm$0.044 & 0.291$\pm$0.050 & 0.129$\pm$0.047 & 0.090$\pm$0.035 \\ 
NOTEARS (NL) & 0.222$\pm$0.048 & 0.434$\pm$0.052 & 0.293$\pm$0.050 & 0.157$\pm$0.068 & 0.332$\pm$0.061 & 0.228$\pm$0.044 & 0.189$\pm$0.041 \\ 
DAG-GNN      & 0.332$\pm$0.071 & 0.413$\pm$0.059 & 0.354$\pm$0.072 & 0.262$\pm$0.049 & \textbf{0.336$\pm$0.060} & 0.285$\pm$0.051 & 0.257$\pm$0.066 \\ 
\modelName      & \textbf{0.698$\pm$0.052} & 0.588$\pm$0.042 & \textbf{0.635$\pm$0.049} & \textbf{0.417$\pm$0.065} & 0.314$\pm$0.069 & \textbf{0.359$\pm$0.059} & \textbf{0.615$\pm$0.074} \\ 
MMHC         & 0.612$\pm$0.057 & \textbf{0.602$\pm$0.051} & 0.601$\pm$0.046 & 0.355$\pm$0.041 & 0.261$\pm$0.041 & 0.298$\pm$0.058 & 0.471$\pm$0.061 \\
Tabu         & 0.332$\pm$0.042 & 0.461$\pm$0.053 & 0.390$\pm$0.047 & 0.121$\pm$0.052 & 0.198$\pm$0.054 & 0.128$\pm$0.051 & 0.240$\pm$0.050 \\
HillClimb    & 0.291$\pm$0.054 & 0.452$\pm$0.060 & 0.361$\pm$0.051 & 0.134$\pm$0.049 & 0.196$\pm$0.061 & 0.130$\pm$0.050 & 0.221$\pm$0.044 \\
\bottomrule
\end{tabular}
    \caption{Structure learning results for the extended synthetic experiment described in \autoref{tab:synthetic_times}. Each value is the mean and standard error over twenty-four datasets. In general, the results are qualitatively similar to those obtained in the synthetic experiment in the paper (recall \autoref{tab:synthetic_causality}), with \modelName obtaining superior performance compared to the previous and the new baselines. 
Notice that the new baseline MMHC is close to \modelName, being superior in adjacency-precision. However, as shown in \autoref{tab:synthetic_times}, MMHC scales poorly.\label{tab:causality_appendix}}
\end{table*}

\begin{figure}
    \centering
    \includegraphics[scale=0.55]{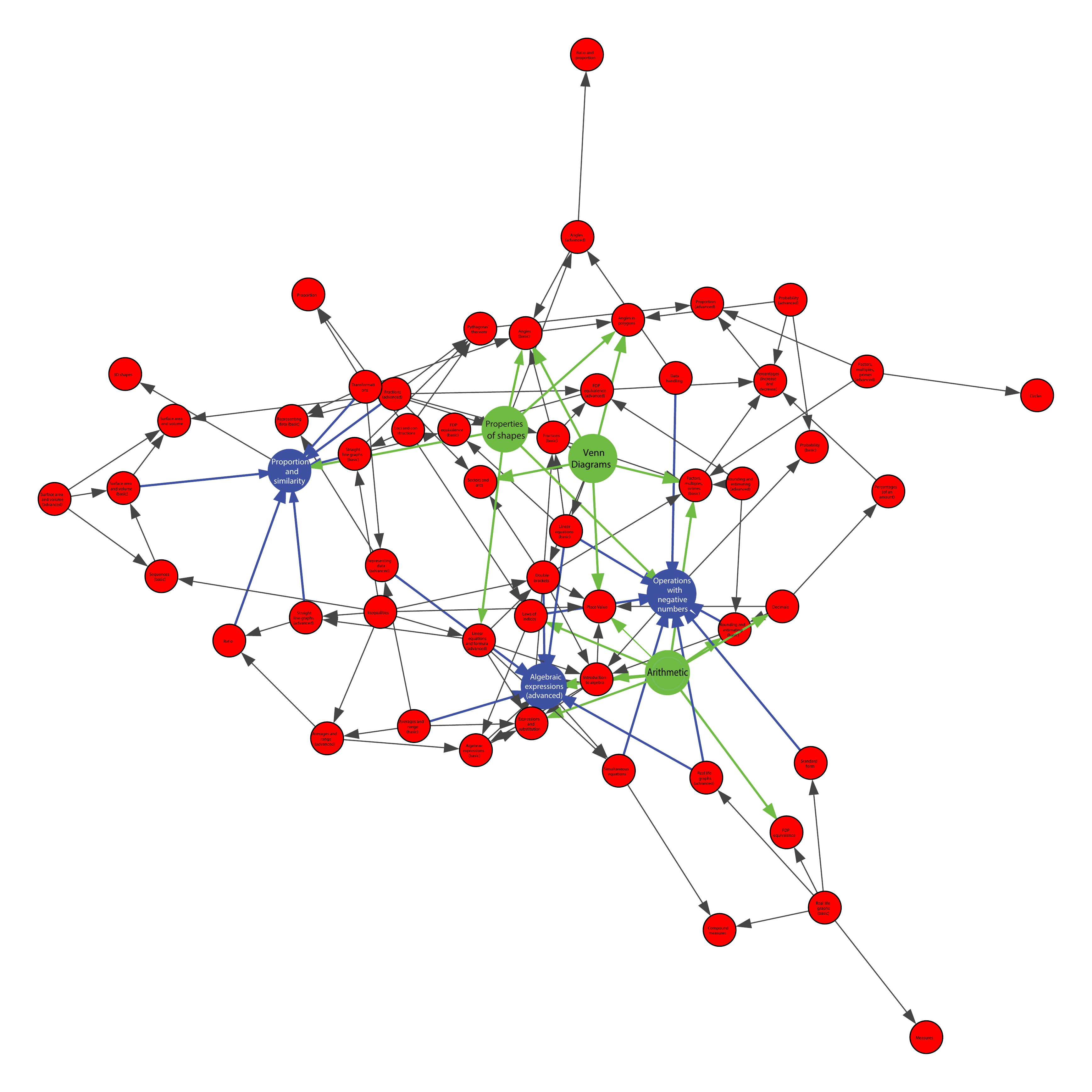}
    \caption{The structures discovered by \modelName based on an alternative \cz{education} dataset. In particular, the \textcolor{myblue}{blue node} represents the curriculum that should be taught later. On the other hand, the \textcolor{mygreen}{green node} represents the fundamental topics that should be taught earlier.}
    \label{fig:eedi_topic_relationship}
\end{figure}